\documentclass[10pt,twocolumn,letterpaper]{article}

\usepackage{cvpr}              

\usepackage{xr}
\usepackage{epsfig}
\usepackage{graphicx,comment}
\usepackage{amsmath}

\usepackage{indentfirst}
\usepackage{algorithm}
\usepackage{algpseudocode} 
\usepackage{amssymb}
\usepackage{verbatim}
\usepackage{multirow}
\usepackage{booktabs}
\usepackage{longtable}
\usepackage{soul}
\usepackage{arydshln}
\usepackage{enumitem}
\usepackage[usenames,dvipsnames,table]{xcolor}

\usepackage{float}
\usepackage{xspace}
\usepackage{bm}
\usepackage[titletoc]{appendix}
\usepackage[resetlabels]{multibib}
\usepackage{capt-of}

\usepackage{sidecap}

\newcites{sec}{References}
 
\newcommand{\SystemName}{GAC-FAS}

\usepackage{upgreek}

\usepackage{color, colortbl}
\definecolor{codegreen}{rgb}{0,0.74,0.38}
\definecolor{codegray}{rgb}{0.5,0.5,0.5}
\definecolor{codepurple}{rgb}{0.58,0,0.82}
\definecolor{backcolour}{RGB}{230, 230, 230}
\definecolor{LightCyan}{rgb}{0.88,1,1}
\definecolor{LightRed}{RGB}{255, 204, 203}

\usepackage{stackengine}
\def\delequal{\mathrel{\ensurestackMath{\stackon[1pt]{=}{\scriptstyle\Delta}}}}

\usepackage{amsmath}
\DeclareMathOperator*{\argmax}{arg\,max}

\newcommand\Tstrut{\rule{0pt}{2.6ex}}       
\newcommand\VRule[1][\arrayrulewidth]{\vrule width #1}


\makeatletter
\def\adl@drawiv#1#2#3{%
        \hskip.5\tabcolsep
        \xleaders#3{#2.5\@tempdimb #1{1}#2.5\@tempdimb}%
                #2\z@ plus1fil minus1fil\relax
        \hskip.5\tabcolsep}
\newcommand{\cdashlinelr}[1]{%
  \noalign{\vskip\aboverulesep
           \global\let\@dashdrawstore\adl@draw
           \global\let\adl@draw\adl@drawiv}
  \cdashline{#1}
  \noalign{\global\let\adl@draw\@dashdrawstore
           \vskip\belowrulesep}}
\makeatother

\usepackage{pifont}
\newcommand{\xmark}{\ding{55}}%

\usepackage[pagebackref,breaklinks,colorlinks]{hyperref}
\hypersetup{
    colorlinks=true, 
    urlcolor=blue,  
}
\usepackage[capitalize]{cleveref}
\crefname{section}{Sec.}{Secs.}
\Crefname{section}{Section}{Sections}
\Crefname{table}{Table}{Tables}
\crefname{table}{Tab.}{Tabs.}

\begin{document}
\title{Gradient Alignment for Cross-Domain Face Anti-Spoofing}
\author{Binh M. Le \quad\quad\quad\quad Simon S. Woo\thanks{Corresponding author.} \\Sungkyunkwan University, South Korea\\{\tt\small \{bmle,swoo\}@g.skku.edu}
}
\maketitle

\begin{abstract}
Recent advancements in domain generalization (DG) for face anti-spoofing (FAS) have garnered considerable attention. Traditional methods have focused on designing learning objectives and additional modules to isolate domain-specific features while retaining domain-invariant characteristics in their representations.  However, such approaches often lack guarantees of consistent maintenance of domain-invariant features or the complete removal of domain-specific features. Furthermore, most prior works of DG for FAS do not ensure convergence to a local flat minimum, which has been shown to be advantageous for DG. In this paper, we introduce \SystemName, a novel learning objective that encourages the model to converge towards an optimal flat minimum without necessitating additional learning modules. Unlike conventional sharpness-aware minimizers, \SystemName\ identifies ascending points for each domain and regulates the generalization gradient updates at these points to align coherently with empirical risk minimization (ERM) gradient updates. This unique approach specifically guides the model to be robust against domain shifts. We demonstrate the efficacy of \SystemName\ through rigorous testing on challenging cross-domain FAS datasets, where it establishes state-of-the-art performance. {The code is available at:  {\url{https://github.com/leminhbinh0209/CVPR24-FAS}}}.

\end{abstract}
\section{Introduction}
\begin{figure}[t]
\centering
\includegraphics[width=8.4cm]{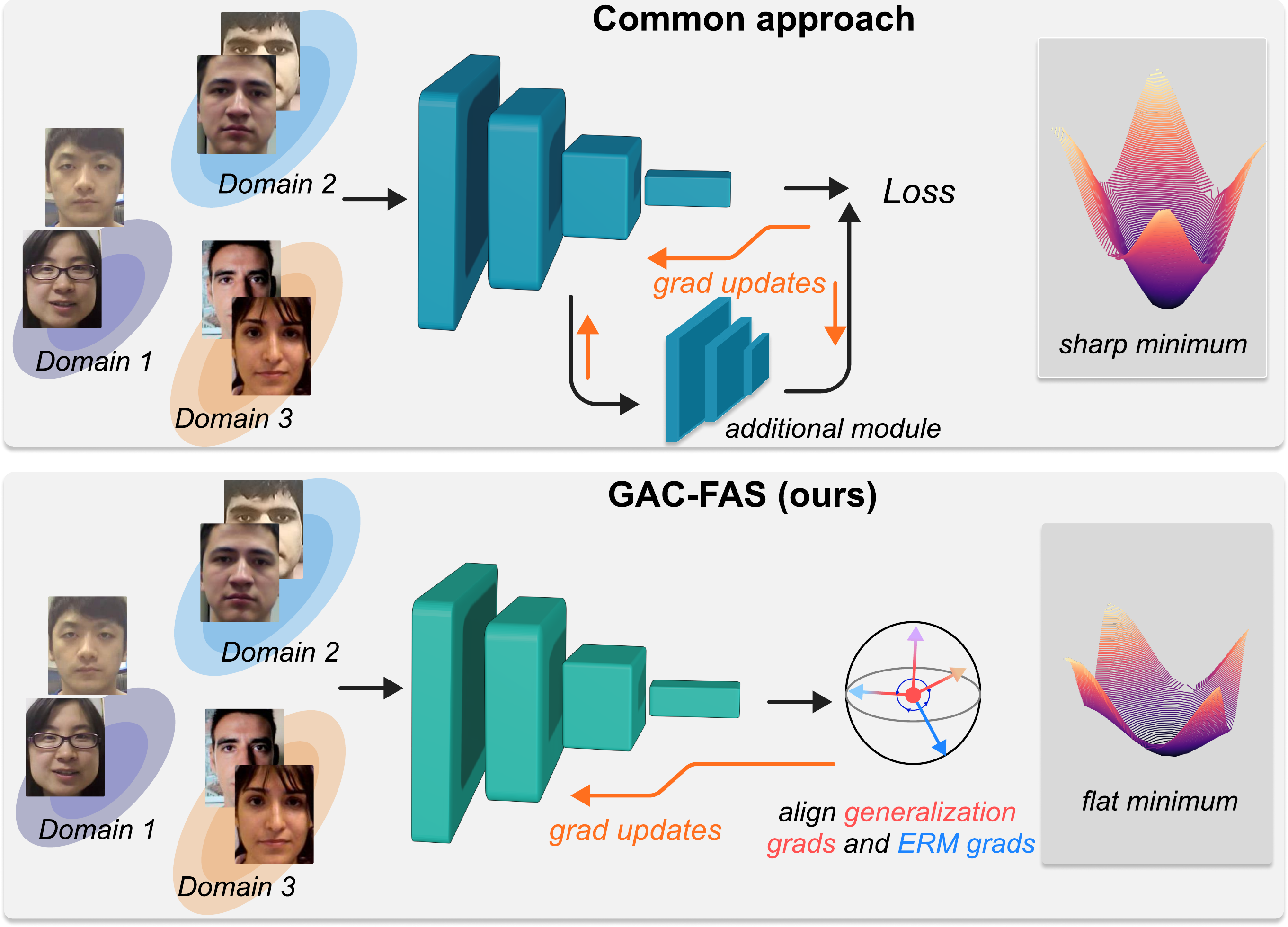}
\caption{\textbf{Illustration of Our Learning Objective.} Most SoTA methods \cite{zhou2023iadg, wang2022ssan, liu2023udgfasssdg} for DG in FAS rely on auxiliary modules to learn domain-invariant features, and do not guarantee convergence towards a flat minimum. In contrast, our method coherently aligns the generalization gradients at ascending points of each domain with gradients derived from ERM. This approach ensures that the model converges to an optimal flat minimum and is robust against domain shifts.}
\label{fig:thumbnail_motivation}
\vspace{-10pt}
\end{figure}
With the increasing importance of security systems, face recognition technologies \cite{deng2019arcface, wang2018cosface, liu2017sphereface} have become ubiquitous in many industrial applications. However, these systems are vulnerable to presentation attacks, such as printed faces \cite{fool2d}, and 3D masks \cite{fool3d}, among others. Consequently, face anti-spoofing (FAS) has emerged as an essential technology to safeguard recognition systems over the past decades \cite{de2013lbp, yang2013face, pan2007eyeblink, kollreider2007real, chetty2010biometric, feng2016integration, lin2019face}. Although existing methods have achieved promising performance, they often suffer from poor generalization when exposed to unseen environments. This limitation is largely due to their assumption of stationary settings, such as lighting conditions or sensor variations, which often do not hold in real-world scenarios.

To address this challenge, recent studies have focused on improving domain generalization (DG) for FAS  by learning domain-invariant features from source training domains \cite{wang2022ssan, jia2020ssgd, wang2022patchnet, sun2023safas, liu2023udgfasssdg, zhou2023iadg, chen2021d2am, shao2020rfm, wang2021sda}. Primary efforts include removing domain-specific features from representations through adversarial training \cite{wang2019improving, shao2019maddg, jia2020ssgd} or meta-learning \cite{chen2021d2am, shao2020rfm, wang2021sda}. Subsequent works have applied metric learning methods \cite{wang2022patchnet, sun2023safas} and style ensemble techniques \cite{wang2022ssan, zhou2023iadg} to enhance robustness under domain shifts. However, most of these studies assume that domain-invariant features are preserved for DG through their specific designs of additional learning modules, without ensuring convergence of the model to a local flat minimum.

In this research, we introduce a novel training objective, namely {Gradient Alignment for Cross-Domain Face Anti-Spoofing} (\SystemName), designed to guide detectors towards an optimal flat minimum robust towards domain shift. This approach is particularly motivated by the recent advancements in Sharpness-Aware Minimization (SAM) \cite{foret2020sharpness, kwon2021asam, zhuang2021gsam}, which offers a promising alternative to empirical risk minimization (ERM)\cite{vapnik1991erm}  for seeking generalizable minima. Our objective function for DG in FAS is carefully modulated by considering the limitations of current SAM variants. When SAM is applied to entire datasets, it may produce biased updates due to the dominance of a particular domain or generate inconsistent gradients when applied to individual domains. Moreover, the updates to the SAM generalization gradient have a tendency to yield a model that is capable of handling many forms of noise, including label noise and adversarial noise. However, our primary focus is on addressing domain shifts in the context of DG for  FAS.

Consequently, we propose two essential conditions for DG in FAS. First, the objective should aim for an optimal flat minimum that is both flat and low in terms of the training loss. Second, the SAM generalization gradient updates, derived at ascending points (see its definition in Sec. \ref{subsec:sam}) for each domain, should be coherently aligned with each other and with the ERM gradient update as illustrated in Fig. \ref{fig:thumbnail_motivation}. This dual approach enables our model to learn a more stable local minimum and become more robust to domain shifts {over different face spoofing datasets}.

Our comprehensive experiments on benchmark datasets under various settings, including leave-one-out, limited source domain, and performance upon convergence, demonstrate the superiority of our method compared to current state-of-the-art (SoTA) baselines. The main contributions of our work are summarized as follows:
\textbf{1)} We offer a new perspective for cross-domain FAS, shifting the focus from learning domain-invariant features to finding an optimal flat minimum for {significantly improving the generalization and} robustness to domain shifts.

\textbf{2)} We propose a novel training objective: self-regulating generalization gradient updates at ascending points to coherently align with the ERM gradient update, benefiting DG in FAS.

\textbf{3)} We demonstrate that our approach outperforms well-known baselines in both snapshot and convergence performance across popular FAS evaluation protocol settings.

Our paper is structured as follows: Sec. \ref{sec:relatedwork} provides a concise review of the most pertinent literature in FAS, along with the relationship between loss landscape sharpness and model generalization. Sec. \ref{sec:method} introduces the preliminaries of ERM and SAM within the context of FAS, followed by a detailed presentation of our approach. Experimental results and ablation studies are discussed in Sec. \ref{sec:results}. Finally, we draw our conclusions in Sec. \ref{sec:conclusion}

\section{Related Work}
\label{sec:relatedwork}

\subsection{Face Anti-Spoofing}
In the initial phases of research, handcrafted features were primarily employed as artifacts for detection. Such features include LBP \cite{de2013lbp, boulkenafet2015face}, HOG \cite{yang2013face, komulainen2013context}, and SIFT \cite{patel2016secure}. Concurrently, studies have examined predefined biometric traits and behaviors, such as eye blinking \cite{pan2007eyeblink}, lip motion \cite{kollreider2007real}, head turning, and facial expression variations \cite{chetty2010biometric}. With the advent of deep neural networks, there has been a notable enhancement in detection capabilities \cite{feng2016integration, li2016original, patel2016cross}. Such improvements were further facilitated through diverse supervisory inputs, encompassing depth maps \cite{yu2021revisiting}, reflection maps \cite{zhang2021structure}, and R-PPG signals \cite{lin2019face}. Recently, transformer-based models have emerged, demonstrating superior efficacy in identifying spoofing attempts \cite{huang2022adaptive, liao2023domain}.

Lately, there has been a growing interest in model generalization across disparate domains. A significant body of work has employed domain adaptation (DA) techniques, where pre-trained models are fine-tuned to novel domains using additional data \cite{li2018unsupervised,wang2020unsupervised,guo2022multi,zhou2022generative}. Concurrently, domain generalization (DG) methodologies, particularly those incorporating adversarial loss, have sought to achieve generalization by extracting domain-invariant features from source training domains \cite{wang2022ssan, jia2020ssgd, wang2022patchnet, sun2023safas, liu2023udgfasssdg, zhou2023iadg}.  Additionally, {several research works} have considered meta-learning as a form of regularization to counteract domain shifts during the training phase \cite{chen2021d2am, shao2020rfm, wang2021sda}, and others have pursued self-supervised learning to reduce reliance on labeled data \cite{liu2021taming, liu2023towards}.

In contrast to prior studies in DG that primarily centered on creating auxiliary modules to eliminate domain-specific features, these approaches may not generalize effectively to unseen domains due to uncertainties  in training model convergence to flat loss regions. {On the other hand,} our approach leverages the inherent sharpness of a model within specific domains by aligning these models. Finally, our goal is to construct a more robust and universally generalizable model for FAS.

\subsection{Sharpness and Generalization}
The relationship between sharpness and model generalization was initially broached in \cite{hochreiter1994simplifying}. Building on this foundation and under the \textit{i.i.d} assumption, numerous theoretical and empirical investigations delved into the relationship from the lens of loss surface geometry \cite{keskar2016large, dziugaite2017computing, garipov2018loss, izmailov2018averaging, foret2020sharpness, cha2021swad}. Notably, both stochastic weight averaging (SWA) \cite{izmailov2018averaging} and stochastic weight averaging densely (SWAD) \cite{cha2021swad} have posited, both theoretically and practically, that a flatter minimum can narrow the DG gap, leading them to propose distinct weight averaging methodologies. Nevertheless, these strategies did not explicitly encourage the model to converge  towards flatter minima during its training phase.

Concurrently, Sharpness-Aware Minimization (SAM) \cite{foret2020sharpness} and its subsequent variants \cite{du2021esam, zhuang2021gsam, liu2022looksam, wang2023sagm} aimed to address the sharp minima problem by adjusting the objective to minimize a perturbed loss, $\mathcal{L}_p(\theta)$, which is defined as the maximum loss within a neighborhood parameter space. Specifically, Look-SAM \cite{liu2022looksam} and ESAM \cite{du2021esam} reduced the computational demands of SAM. However, they retained SAM's primary challenge, wherein the perturbed loss $\mathcal{L}_p(\theta)$ could potentially disagree with the actual sharpness measure. To overcome this challenge, {GSAM} \cite{zhuang2021gsam} minimized a surrogate gap, $h(\theta)\delequal\mathcal{L}_p(\theta)-\mathcal{L}(\theta)$, albeit at the expense of increasing $\mathcal{L}(\theta)$. Later on, SAGM \cite{wang2023sagm} rectified the inconsistencies observed in GSAM by incorporating gradient matching, ensuring model convergence to flatter regions. While SAM-based techniques have shown promise in generalizing from a single source and dealing with various types of noise, including label and adversarial noise, their application in multi-source domain DG for FAS has not been explored. Inspired by these foundational studies, we have tailored SAM to ensure that its generalization gradient updates derived from multi-source domains are aligned with each another, and {with} the ERM gradient update.  To the best of our {knowledge}, this study pioneers the exploration of SAM's capacity for DG in FAS.

\section{Methods}
In this section, we first define the general empirical risk for training cross-domain FAS problems. Next, we revisit variations of Sharpness-Aware Minimization for domain generalization from which we draw our motivation. Furthermore, we propose our approach, \SystemName, specifically tailored for the problem of DG for FAS. Finally, we analyze the benefits and prove the convergence rate of our algorithm.
\label{sec:method}
\subsection{ Problem Definition}
\label{subsec:preliminaries}
We begin by {introducing} the notion of cross-domain FAS. Consider an input space $\boldsymbol{\mathcal{X}}\in \mathbb{R}^{d}$ and an output space $\boldsymbol{\mathcal{Y}}=\{0 \text{ (fake or {spoofed})}, 1 \text{ (live)}\}$. Assuming there are $k$ distinct source domains for training, represented as $\boldsymbol{\mathcal{S}}=\{ \textcolor{blue}{\mathcal{S}_{i}}\}^k_i$, and a singular target domain denoted by $\boldsymbol{\mathcal{T}}$.

A neural network, characterized as $f:\boldsymbol{\mathcal{X}}\rightarrow \boldsymbol{\mathcal{Y}}$, is parameterized by learning parameters $\theta$. Its aim is to distinguish whether an input $x$ from the source domains is live or spoofed {(fake)}. A standard approach to optimization involves the empirical risk minimization (ERM) framework \cite{vapnik1991erm}, which aims to minimize the loss described by:
\begin{align}
     \min_{\theta}\mathcal{L}(\theta;\boldsymbol{\mathcal{S}})&=\min_{\theta}\mathbb{E}_{\textcolor{blue}{\mathcal{S}_{i}}\sim\boldsymbol{\mathcal{S}}}\mathcal{L}(\theta;\textcolor{blue}{\mathcal{S}_{i}}),
    \label{eqn:erm}
\end{align}
where $\mathcal{L}(\theta;\textcolor{blue}{\mathcal{S}_{i}})=\mathbb{E}_{(x,y)\sim\textcolor{blue}{\mathcal{S}_{i}}}\ell(f(x;\theta),y)$ is domain-wise empirical loss on domain $i-th$, and $\ell$ could be cross-entropy loss \cite{jia2020ssgd} or $L_1$ regression loss \cite{george2019deep}.

To minimize the empirical risk $\mathcal{L}(\theta;\boldsymbol{\mathcal{S}})$, the neural network $f$ aspires to identify the optimal parameter set $\theta^*$. A notable challenge with ERM is its propensity to overfit the training data and converge towards sharp minima, compromising the performance on an unseen domain. Such tendencies might arise due to domain-specific attributes {such as} camera configurations or image resolution \cite{sun2023safas}. Consequently, targeting flatter minima when training on source domains becomes pivotal for addressing the DG for FAS problem.

\subsection{Preliminaries: Sharpness-Aware Minimization}
\label{subsec:sam}
The Sharpness-Aware Minimization (SAM) \cite{foret2020sharpness} method aims to identify a flatter area in the vicinity of the minimum that exhibits a lower loss value. In order to attain this objective, given a training set $\mathcal{D}$ (which can be considered as either $\boldsymbol{\mathcal{S}}$ or $\textcolor{blue}{\mathcal{S}_{i}}$ in our problem later), SAM addresses the subsequent min-max problem:
\begin{align}
    &\min_{\theta}\mathcal{L}_{p}(\theta;\mathcal{D}) + \mathcal{R}(\theta;\mathcal{D}),\text{ where }\\
    &\mathcal{L}_{p}(\theta;\mathcal{D})=\max_{\epsilon \in \mathcal{B}(\theta,\rho)}\mathcal{L}(\theta+\epsilon;\mathcal{D}),
\label{eq:sam_eqn}
\end{align}
and $\mathcal{R}(\theta;\mathcal{D})$ is a regularization term, and $\mathcal{B}(\theta,\rho)=\{\epsilon : \lVert\epsilon-\theta\rVert\leq \rho \}$ is the vicinity of model weight vector $\theta$ with a predefined constant radius $\rho$. Intuitively,  for a given $\theta$, the maximization in Eq.  \ref{eq:sam_eqn} identifies the most adversarial weight perturbation, denoted as $\epsilon^*$, within the ball $\mathcal{B}$ of radius $\rho$. This perturbation maximizes the empirical loss, leading to $\mathcal{L}(\theta+\epsilon^*;\mathcal{D})$ being the supremum in $\mathcal{B}(\theta,\rho)$. We now refer $\epsilon^*$ and $\theta+\epsilon^*$ as  \underline{\textbf{ascending vector}} and \underline{\textbf{ascending point}}, respectively. By minimizing $\mathcal{L}(\theta+\epsilon^*;\mathcal{D})$, the approach encourages the selection of $\theta$ values that are situated in a region with a flatter loss landscape. Consequently, the function $f$ exhibits enhanced stability under domain shifts, making it more resilient to unseen domains.

SAM uses Taylor expansion of the empirical loss  around $\theta$ to estimate $\epsilon^*$ as follows \cite{foret2020sharpness}:
\begin{equation}
    \hat{\epsilon}=\rho \frac{\nabla\mathcal{L}(\theta;\mathcal{D})}{\lVert \nabla \mathcal{L}(\theta;\mathcal{D}) \rVert}\approx\argmax_{\epsilon \in \mathcal{B}(\theta,\rho)}\mathcal{L}(\theta+\epsilon;\mathcal{D}).
\label{eq:est_eps}
\end{equation}
Therefore, the perturbation loss of SAM reduces to 
\begin{equation}
    \mathcal{L}_{p}(\theta;\mathcal{D})=\mathcal{L}(\theta+\hat{\epsilon};\mathcal{D}), \text{ where  } \hat{\epsilon}=\rho\frac{\nabla\mathcal{L}(\theta;\mathcal{D}) }{\lVert \nabla \mathcal{L}(\theta;\mathcal{D}) \rVert}.
    \label{eqn:sam_est}
\end{equation}
\begin{figure*}[t]
\centering
\includegraphics[width=17cm]{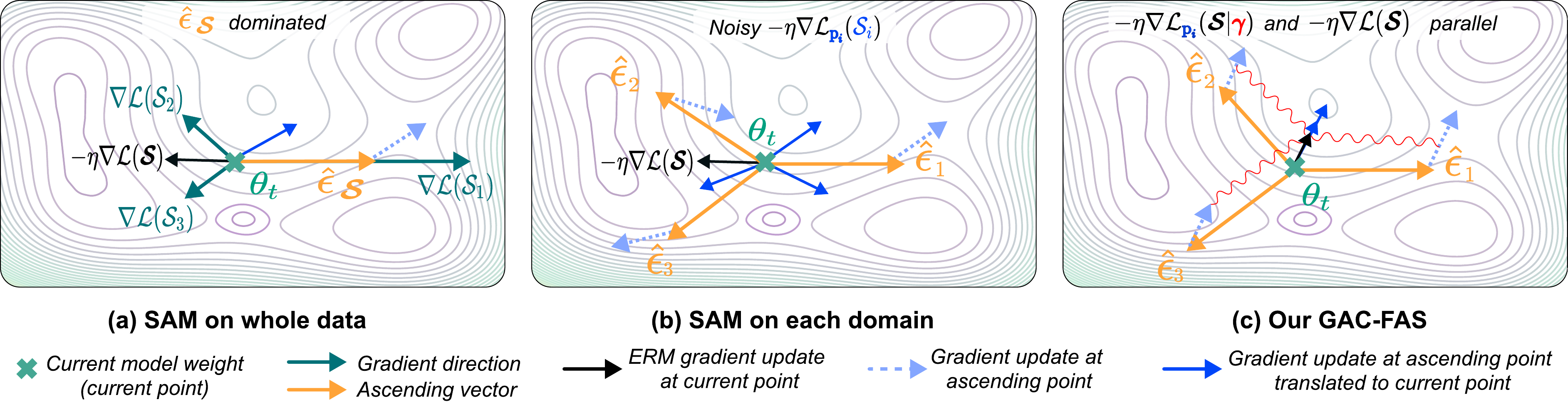}
\caption{ \textbf{Illustration of Different SAM Objective Approach.} (a) Standard SAM variants applied to the entire source dataset yield a biased ascending vector $\hat{\epsilon}_{\boldsymbol{\mathcal{S}}}$, predominantly influenced by a particular domain. (b) Domain-specific gradient adjustments lead to noisy gradient estimates, impeding optimization progress. (c) Our proposed \SystemName\ addresses these issues by computing perturbation losses across the dataset at all ascending points, while concurrently adjusting gradients to align with the ERM gradients at the current point (with  $\gamma$), making the model robust to domain shift.
 }
\label{fig:objective}

\end{figure*}



Subsequently,  Zhuang \textit{et al.} \cite{zhuang2021gsam} and  Wang \textit{et al.} \cite{wang2023sagm} introduce a  surrogate gap (sharpness)
$h(\theta)\delequal\mathcal{L}_p(\theta)-\mathcal{L}(\theta)$, albeit at the expense of increasing $\mathcal{L}(\theta)$
into their objectives to obtain a better minimum. Nevertheless, there are the following limitations and  issues when applying SAM and its variants for DG for FAS, as discussed subsequently.


\noindent \textbf{Our Preliminary Observations {and Analysis}.} (\textit{i})  \textbf{Compensation for Input Changes}: Optimizing $\mathcal{L}_{p}(\theta;\mathcal{D})$ acts as a {compensatory}  mechanism for a range of input changes. These include domain shifts and various types of corruptions such as adversarial noises \cite{wu2020adversarial}, heavy compression \cite{le2023quality}, and label noises \cite{foret2020sharpness}. However, our primary objective in this research is to enhance robustness against domain shifts specific to FAS. 
(\textit{ii}) \textbf{Dominated $\hat{\epsilon}$ in All-Domain Application ($\mathcal{D}=\boldsymbol{\mathcal{S}}$)}: By applying Eq.  \ref{eqn:sam_est} or its variants to the problem of DG for FAS as formulated in Eq.~\ref{eqn:erm}, we can determine the optimal ascending vector $\hat{\epsilon}$  for the training source dataset $\mathcal{D}=\boldsymbol{\mathcal{S}}$. However, a significant challenge arises from imbalances in dataset sizes and the presence of subtle, domain-specific artifacts. This can lead to `learning shortcuts,' where the model preferentially learns from less complex or larger domains. Consequently,  the ascending vector $\hat{\epsilon}$ from ($\theta$, $\boldsymbol{\mathcal{S}}$) can become dominated by a particular domain $\mathcal{S}_i$. This behaviours is similar with the long-tail problem \cite{zhou2023imbsam, zhou2023class}.
(\textit{iii}) \textbf{Gradient Conflicts in Domain-wise Application ($\mathcal{D}=\textcolor{blue}{\mathcal{S}_{i}}$)}: On the other hand, if we apply  Eq.  \ref{eqn:sam_est} or its variants on domain-wise manner, \textit{i.e.}, $\mathcal{D}=\textcolor{blue}{\mathcal{S}_{i}}$, the gradients of the model, derived from each domain as their ascending points, can counteract one another, leading to potential conflicts between domains. This phenomenon is further illustrated in Fig.~\ref{fig:objective}.
In light of these intricacies and {limitations}, we argue that the SAM variants might not deliver optimal generalization performance for the DG for FAS problem.

\subsection{Objective of \SystemName}
\label{subsec:obj}
Given the insights gleaned from our prior analysis, we {introduce} two pivotal conditions to ensure that our model remains robust across unseen domains: (i)  \textbf{Optimal minimum}: The identified minimum should not only be sufficiently low but should also reside on a flat loss surface. (ii) \textbf{Aligned cross-domain gradients}: From the source training datasets, the generalization gradient update learned from some domains should align with the ERM gradient of another domain. 

Intuitively, the first condition resonates with the core objectives of optimal minimum of loss landscape from training set, as discussed previously\cite{zhuang2021gsam, wang2023sagm} .
The second condition serves dual purposes. It aims to harmonize the optimization of $h(\theta)$, reducing potential conflicts. Simultaneously, it aspires for $\mathcal{L}_{p}(\theta;\boldsymbol{\mathcal{S}})$ to exclusively compensate for domain shifts in FAS.

To fulfill these conditions, we introduce a novel optimization objective for our DA FAS, expressed as:
\begin{equation}
 \mathcal{L}(\theta;\boldsymbol{\mathcal{S}})+\mathbb{E}_{\textcolor{blue}{\mathcal{S}_{i}}\sim\boldsymbol{\mathcal{S}}}\mathcal{L}_{\textcolor{blue}{p_i}}(\theta-\gamma\nabla \mathcal{L}(\theta;  \boldsymbol{\mathcal{S}});\boldsymbol{\mathcal{S}})+\mathcal{R}(\theta;\boldsymbol{\mathcal{S}}).
\label{eqn:gac_fas1}
\end{equation}

In this formulation, which is inspired by  \cite{wang2023sagm}, we perturb the model weights \textit{w.r.t} each individual domain. Simultaneously, we incorporate an auxiliary ERM's gradient term, $\gamma\nabla\mathcal{L}(\theta;\boldsymbol{\mathcal{S}})$, computed over all source training domains.  {And,} Eq.  \ref{eqn:gac_fas1}  can be further {expressed} as:
\begin{equation}
  \mathcal{L}(\theta;\boldsymbol{\mathcal{S}})+\mathbb{E}_{\textcolor{blue}{\mathcal{S}_{i}}\sim\boldsymbol{\mathcal{S}}}\mathcal{L}(\theta+\textcolor{blue}{\hat{\epsilon}_i} - \gamma\nabla \mathcal{L}(\theta;  \boldsymbol{\mathcal{S}});\boldsymbol{\mathcal{S}})+\mathcal{R}(\theta;\boldsymbol{\mathcal{S}}),
\label{eqn:gac_fas2}
\end{equation}
where the optimal perturbation is characterized by $ \textcolor{blue}{\hat{\epsilon}_i}= \rho\frac{\nabla\mathcal{L}(\theta;\textcolor{blue}{\mathcal{S}_{i}}) }{\lVert \nabla \mathcal{L}(\theta;\textcolor{blue}{\mathcal{S}_{i}}) \rVert}$ as defined in Eq.  \ref{eq:est_eps}.

\subsection{Benefits \& Convergence of \SystemName}
\label{subsec:analysis}
In this section, we offer analyses for a deeper understanding of our proposed losses for DA in FAS, detailing how they satisfy the two conditions outlined in Sec. \ref{subsec:obj}. Subsequently, we present our novel training algorithm and the theorem regarding its convergence rate.

\begin{figure}[t]
\centering
\includegraphics[width=8.0cm]{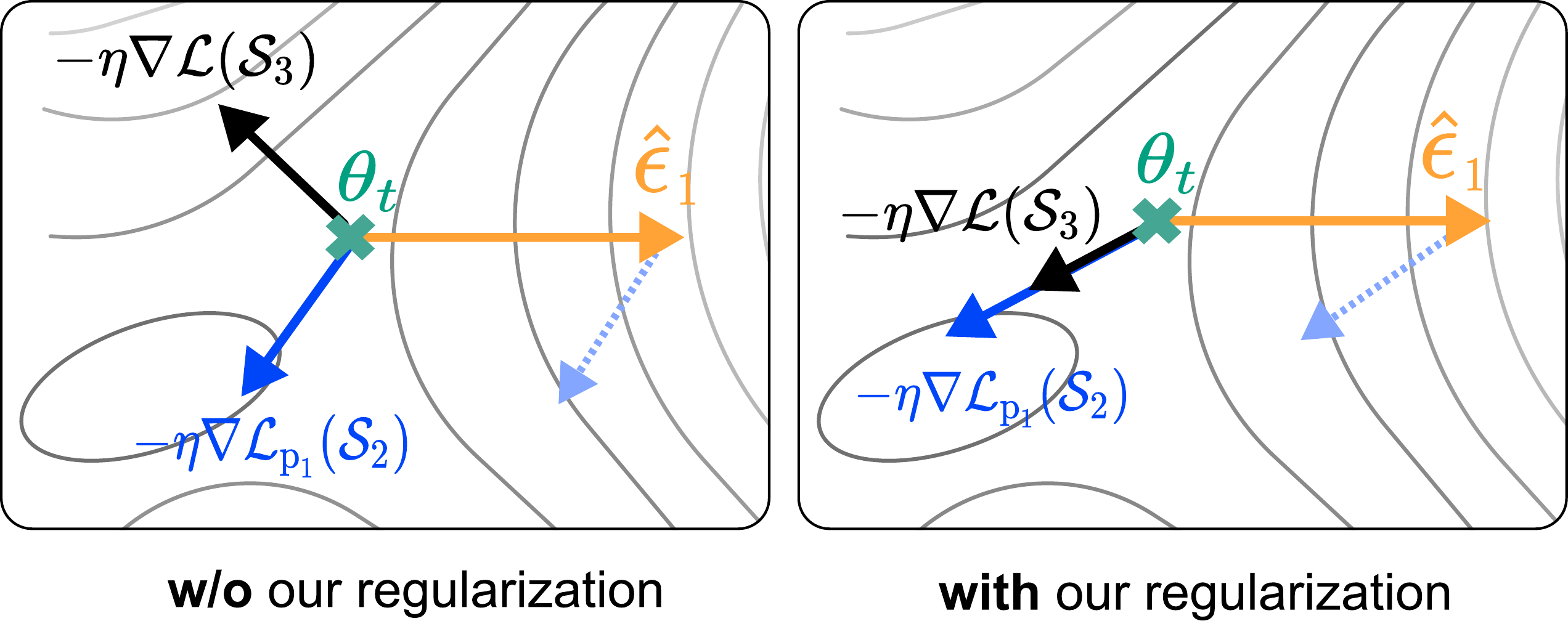}
\caption{ Illustration of the effects described by Eq. \ref{eqn:sum_sum}. The term $-\eta\nabla \mathcal{L}_{\text{p}_1}(\mathcal{S}_2)$ represents the generalization gradient update of the model learned from $\mathcal{S}_1$ and $\mathcal{S}_2$. The term $-\eta\nabla\mathcal{L} (\mathcal{S}_3)$ denotes the ERM update for domain $\mathcal{S}_3$ and serves as a comparative oracle for domain shift. In the absence of our regularization, the generalization update is not robust to the domain shift associated with $\mathcal{S}_3$ (left), as their update directions are different. Conversely, with our regularization as formulated in Eq. \ref{eqn:sum_sum}, the generalization update aligns with the ERM update on $\mathcal{S}_3$, suggesting that the model updates in a direction that is robust to domain shifts (right). }
\label{fig:effect}
\vspace{-10pt}
\end{figure}

\subsubsection{Benefits of \SystemName}
We perform the first order Taylor expansion around $\theta+\textcolor{blue}{\hat{\epsilon}_i}$ for the second term in Eq.  \ref{eqn:gac_fas1} as follows:
\begin{align}
    &\mathbb{E}_{\textcolor{blue}{\mathcal{S}_{i}}\sim\boldsymbol{\mathcal{S}}}\mathcal{L}_{\textcolor{blue}{p_i}}(\theta-\gamma\nabla\mathcal{L}(\theta;\boldsymbol{\mathcal{S}});\boldsymbol{\mathcal{S}}) \nonumber\\
    \approx &\mathbb{E}_{\textcolor{blue}{\mathcal{S}_{i}}\sim\boldsymbol{\mathcal{S}}}\mathcal{L}_{\textcolor{blue}{p_i}}(\theta;\boldsymbol{\mathcal{S}}) -\gamma \langle\nabla\mathcal{L}_{\textcolor{blue}{p_i}}(\theta;\boldsymbol{\mathcal{S}}),\nabla\mathcal{L}(\theta;\boldsymbol{\mathcal{S}}) \rangle .      
\label{eqn:taylor_gac_fas}
\end{align}
As a result, our objective in Eq.  \ref{eqn:gac_fas1} can be {expressed} as follows:
\begin{align}
     &\mathcal{L}(\theta;\boldsymbol{\mathcal{S}})+ \mathbb{E}_{\textcolor{blue}{\mathcal{S}_{i}}\sim\boldsymbol{\mathcal{S}}}\mathcal{L}_{\textcolor{blue}{p_i}}(\theta;\boldsymbol{\mathcal{S}}) \nonumber \\&-\gamma \langle\nabla\mathcal{L}_{\textcolor{blue}{p_i}}(\theta;\boldsymbol{\mathcal{S}}),\nabla\mathcal{L}(\theta;\boldsymbol{\mathcal{S}}) \rangle  +\mathcal{R}(\theta;\boldsymbol{\mathcal{S}}).   
\label{eqn:expand_gac_fas}
\end{align}
The underlying objectives of our Eq. \ref{eqn:expand_gac_fas} is twofold: we aim to minimize the loss functions, namely $\mathcal{L}(\theta;\boldsymbol{\mathcal{S}})$ and $\mathcal{L}_{\textcolor{blue}{p_i}}(\theta;\boldsymbol{\mathcal{S}})$, while simultaneously maximizing the inner products between the gradient of $\mathcal{L}_{\textcolor{blue}{p_i}}(\theta;\boldsymbol{\mathcal{S}})$  and the gradients $\mathcal{L}(\theta;\boldsymbol{\mathcal{S}})$.
 Specifically, condition (\textit{i}) is satisfied by minimizing  $\mathcal{L}(\theta;\boldsymbol{\mathcal{S}})$ and $\mathcal{L}_{\textcolor{blue}{p_i}}(\theta;\boldsymbol{\mathcal{S}}), \forall i$. Our distinct contribution, however, lies in our method's emphasis on cross-domain gradient alignment. This facet of our methodology particularly benefits the DG for FAS as elucidated  subsequently.
\begin{algorithm}[t!]
\caption{Training pipeline for \SystemName.}
\begin{algorithmic}[1]
\Require   DNN $f$
parameterized by $\theta$, training dataset $\boldsymbol{\mathcal{S}}=\{ \textcolor{blue}{\mathcal{S}_{i}}\}^k_i$. Learning rate $\eta$. Alignment parameter $\gamma$ and radius $\rho$. Total number of iterations $T$.
    \For{$t \leftarrow 1$ \textbf{to} $T$}
        \State  Sample a mini-batch $\mathcal{B}=\mathcal{B}_{\mathcal{S}_1}+...+\mathcal{B}_{\mathcal{S}_k}$;
        \State Compute grad. of reg. term $\nabla\mathcal{R}(\theta_t;\mathcal{B})$;
        \State \texttt{\#\small{Compute grad. for $1^{st}$ term of Eq.~\ref{eqn:gac_fas2}:}}
        \State Compute the training loss gradient on each domain $\{\nabla\mathcal{L}(\theta_t;\mathcal{B}_{\textcolor{blue}{\mathcal{S}_{i}}})\}_{i=1}^{k}$, and sum them up to obtain $\nabla\mathcal{L}(\theta_t;\mathcal{B})$;
        
        \State \texttt{\#\small{Compute grad. for $2^{nd}$ term of Eq.~\ref{eqn:gac_fas2}:}}
        \For{ domain $i \in \{1,...,k \}$}
            \State $\textcolor{blue}{\hat{\epsilon}_i}=\rho\frac{\nabla\mathcal{L}(\theta_t;\mathcal{B}_{\textcolor{blue}{\mathcal{S}_{i}}})}{\lVert\nabla\mathcal{L}(\theta_t;\mathcal{B}_{\textcolor{blue}{\mathcal{S}_{i}}})\rVert}$  \texttt{\#\small{ascending vector}}
            \State $\nabla\mathcal{L}_{p}^i=\nabla \mathcal{L}(\theta_{t}+\textcolor{blue}{\hat{\epsilon}_i}-\gamma\nabla\mathcal{L}(\theta_t;\mathcal{B});\mathcal{B})$
        \EndFor
        \State \texttt{\#\small{Update weights:}}
        \State $\theta_{t+1}=\theta_{t} -\eta \cdot \big( \nabla\mathcal{L}(\theta_t;\mathcal{B}) +\frac{1}{k}\Sigma_{i=1}^{k}\nabla\mathcal{L}_{p}^i + \nabla\mathcal{R}(\theta_t;\mathcal{B})\big)$  
        \State $t=t+1$
    \EndFor 
\end{algorithmic}
\label{alg:gac_fas}
\end{algorithm}
By noting that $\nabla\mathcal{L}(\theta;\boldsymbol{\mathcal{S}})=\Sigma_{m=1}^{k}\nabla\mathcal{L}(\theta;{\mathcal{S}_m})$, maximizing $\langle\nabla\mathcal{L}_{\textcolor{blue}{p_i}}(\theta;\boldsymbol{\mathcal{S}}),\nabla\mathcal{L}(\theta;\boldsymbol{\mathcal{S}}) \rangle$ is similar with maximizing following term:
\begin{equation}
\Sigma^k_{m=1}\Sigma^k_{n=1}\langle\nabla\mathcal{L}_{\textcolor{blue}{p_i}}(\theta;{\mathcal{S}_m}),\nabla\mathcal{L}(\theta;{\mathcal{S}_n})\rangle.
\label{eqn:sum_sum}
\end{equation}

We interest in the case where $\mathcal{S}_{n}$ is different with $\mathcal{S}_{m}$ and $\textcolor{blue}{\mathcal{S}_{i}}$. Maximizing the inner product $\langle\nabla\mathcal{L}_{\textcolor{blue}{p_i}}(\theta;{\mathcal{S}_m}), \nabla\mathcal{L}(\theta;{\mathcal{S}_n})\rangle$ implies that the \textit{generalization gradient} update, learned from $\textcolor{blue}{\mathcal{S}_{i}}$ (via $\textcolor{blue}{\hat{\epsilon}_i}$) and $\mathcal{S}_m$, must align with the \textit{ERM gradient} of another domain, $\mathcal{S}_n$. This ERM gradient acts as a comparative oracle for domain shifts, guiding generalization update of the model converge towards a  minimum that is robust against domain shifts.
A toy example of this effect is {illustrated} in Fig.~\ref{fig:effect}. Moreover, maximizing Eq.  \ref{eqn:sum_sum} indirectly leads to the matching each pair of $\{\nabla\mathcal{L}(\theta, \mathcal{S}_n)\}^k$, benefiting for DG \cite{shi2021gradient}.

\subsubsection{Convergence of \SystemName}
\textbf{Theorem 1} (Proof in Sec. \ref{supp:prooftheorem1} Supp. material).  \textit{Suppose that the loss function $\ell(\theta_{t}) = \ell(f(x;\theta_{t}), y)$ satisfies the following assumptions. (i) its gradient  $g(\theta_{t})=\nabla\ell(\theta_{t})$  is  bounded, i.e., $\lVert g(\theta_{t})\rVert \leq G \text{, } \forall t$. (ii) The stochastic gradient is L-Lipchitz, i.e., $\lVert g(\theta_{t})-g(\theta_{t}^\prime)\rVert \leq L\lVert \theta_{t}-\theta_{t}^\prime\rVert$, $\forall \theta_{t}$, $\theta_{t}^\prime$.  Let the learning rate $\eta_t$ be $ \frac{\eta_0}{\sqrt{t}}$, and and let the perturbation be proportional to the learning rate, i.e., $\rho_t=\frac{\rho}{\sqrt{t}}$, and $\gamma_t=\frac{\gamma}{\sqrt{t}}$, we have:}
  \begin{align} 
      &\frac{1}{T}\Sigma^{T}_{t=1}\mathbb{E}_{\textcolor{blue}{\mathcal{S}_{i}}\sim\boldsymbol{\mathcal{S}}}\mathbb{E}_{(x,y)\sim\textcolor{blue}{\mathcal{S}_{i}}}\left[ \lVert \nabla \ell(\theta_{t})\rVert^{2}\right] \leq \mathcal{O}\left(\frac{\log T}{\sqrt{T}}\right), \text{and}  \nonumber\\
      &\frac{1}{T}\Sigma^{T}_{t=1}\mathbb{E}_{\textcolor{blue}{\mathcal{S}_{i}}\sim\boldsymbol{\mathcal{S}}}\mathbb{E}_{(x,y)\sim\textcolor{blue}{\mathcal{S}_{i}}}\left[ \lVert \nabla \ell(\theta_{t}^{\text{adv}})\rVert^{2}\right] \leq \mathcal{O}\left(\frac{\log T}{\sqrt{T}}\right), \nonumber
  \end{align}
 \textit{where  $\theta_{t}^{\text{adv}}=\theta_{t}+\hat{\epsilon_t}-\gamma_t\delta_t$, $\delta_t=\Sigma_{j=1}^{k}\nabla\ell(f(x_j^\prime;\theta_t),y_j^\prime)$, and $(x_j^\prime,y_j^\prime) \sim\mathcal{S}_{j}$. }

Theorem 1 implies that both $\ell$ and $\ell_p$ converge at rate $\mathcal{O}(\log T/\sqrt{T})$, and it matches convergence rate of first-order gradient optimizers such as Adam \cite{kingma2014adam}.

Equipped with Theorem 1, we present the overall training pipeline of our  \SystemName~   in Algorithm \ref{alg:gac_fas}.

\begin{table*}[th!]
\centering
\resizebox{.84\textwidth}{!}{%
    \begin{tabular}{l|ll|ll|ll|ll}
    \hline \multirow{2}{*}{ Methods } & \multicolumn{2}{c |}{\textbf{ICM {$\rightarrow$} O}} & \multicolumn{2}{c |}{ \textbf{OCM {$\rightarrow$} I} } & \multicolumn{2}{c |}{ \textbf{OCI {$\rightarrow$} M} } & \multicolumn{2}{c}{ \textbf{OMI {$\rightarrow$} C} } \\
    & HTER $\downarrow$ & AUC $\uparrow$ & HTER $\downarrow$ & AUC $\uparrow$ & HTER $\downarrow$ & AUC $\uparrow$ & HTER $\downarrow$ & AUC $\uparrow$ \\
    \hline
    \hline
    MMD-AAE \cite{li2018mmdaae} & 40.98 & 63.08 & 31.58 & 75.18 & 27.08 & 83.19 & 44.59 & 58.29 \\
    MADDG \cite{shao2019maddg} & 27.98 & 80.02 & 22.19 & 84.99 & 17.69 & 88.06 & 24.50 & 84.51 \\
    RFM \cite{shao2020rfm} & 16.45 & 91.16 & 17.30 & 90.48 & 13.89 & 93.98 & 20.27 & 88.16 \\
    SSDG-M \cite{jia2020ssgd}  & 25.17 & 81.83 & 18.21 & 94.61 & 16.67 & 90.47 & 23.11 & 85.45 \\
    SSDG-R \cite{jia2020ssgd}  & 15.61 & 91.54 & 11.71 & 96.59 & 7.38 & 97.17 & 10.44 & 95.94 \\
    D$^2$AM \cite{chen2021d2am} & 15.27 & 90.87 & 15.43 & 91.22 & 12.70 & 95.66 & 20.98 & 85.58 \\
    SDA \cite{wang2021sda} & 23.10 & 84.30 & 15.60 & 90.10 & 15.40 & 91.80 & 24.50 & 84.40 \\
    DRDG \cite{liu2021drdg} & 15.63 & 91.75 & 15.56 & 91.79 & 12.43 & 95.81 & 19.05 & 88.79 \\
    ANRL \cite{liu2021anrl} & 15.67 & 91.90 & 16.03 & 91.04 & 10.83 & 96.75 & 17.85 & 89.26 \\
    SSAN \cite{wang2022ssan}& 13.72 &  93.63 & 8.88 & 96.79 & 6.67 & 98.75 & 10.00 & 96.67 \\
    AMEL \cite{zhou2022amel} & 11.31 & 93.96 & 18.60 & 88.79 & 10.23 & 96.62 & 11.88 & 94.39 \\
    EBDG \cite{du2022ebdg} & 15.66 & 92.02 & 18.69 & 92.28 & 9.56 & 97.17 & 18.34 & 90.01 \\
    PathNet \cite{wang2022patchnet} &11.82 &95.07 & 13.40 & 95.67 & 7.10 & 98.46 & 11.33 & 94.58 \\
    IADG \cite{zhou2023iadg} & 8.86 & 97.14 & 10.62 & 94.50 & 5.41 & 98.19 & 8.70 & 96.40 \\
    SA-FAS \cite{sun2023safas} & {10.00} & {96.23} &6.58 &97.54  & {5.95} & 96.55 & {8.78} &95.37\\
     UDG-FAS \cite{liu2023udgfasssdg} &10.97 &95.36 & {5.86} & {98.62} & {5.95} & \textbf{98.47} & 9.82 & \textbf{96.76} \\
    \hline
    \rowcolor{backcolour} \SystemName\ (\textit{\textbf{ours}}) & \textbf{8.60}$^{0.28}$ & \textbf{97.16}$^{0.40}$ & \textbf{4.29}$^{0.83}$  & \textbf{98.87}$^{0.60}$ & \textbf{5.00}$^{0.00}$ & 97.56$^{0.06}$ & \textbf{8.20}$^{0.43}$ &  95.16$^{0.09}$ \\
    \hline
    \hline
    \end{tabular}%
    }
\caption{Evaluation of cross-domain face anti-spoofing on four leading benchmark datasets: CASIA (\textbf{C}), Idiap Replay (\textbf{I}), MSU-MFSD (\textbf{M}), and Oulu-NPU (\textbf{O}). Methods are benchmarked for optimal performance using the standard evaluation procedure outlined in \cite{jia2020ssgd}. Symbols $\uparrow$ and $\downarrow$ signify that larger and smaller values are preferable, respectively.} 
\label{tab:main_results}

\end{table*}
\section{Experimental Results}
\label{sec:results}
In this section, we {compare} our method {with} previous SoTA baselines using standard {FAS evaluation} protocol settings. Additionally, we assess the effectiveness of our algorithm in scenarios where the training source is limited. We then compare its convergence performance with other baselines. Finally, we conduct several ablation studies to explore alternatives to the minimizer for DG in FAS and the effects of hyperparameter tuning.
\begin{table}[t!]
\centering
\resizebox{.48\textwidth}{!}{%
\begin{tabular}{l|ll|ll}
    \hline \multirow{2}{*}{ Methods } & \multicolumn{2}{|c|}{ \textbf{MI $\rightarrow$ C} } & \multicolumn{2}{c}{ \textbf{MI $\rightarrow$ O} } \\
    & HTER $\downarrow$ & AUC $\uparrow$ & HTER $\downarrow$ & AUC $\uparrow$ \\
    \hline \hline
     MSLBP \cite{maatta2011mslpb} & 51.16 & 52.09 & 43.63 & 58.07 \\
Color Texture \cite{boulkenafet2015face}  & 55.17 & 46.89 & 53.31 & 45.16 \\
LBPTOP \cite{de2013lbp} & 45.27 & 54.88 & 47.26 & 50.21 \\
    MADDG \cite{shao2019maddg} & 41.02 & 64.33 & 39.35 & 65.10 \\
    SSDG-M \cite{jia2020ssgd} & 31.89 & 71.29 & 36.01 & 66.88 \\
    D$^2$AM \cite{chen2021d2am} & 32.65 & 72.04 & 27.70 & 75.36 \\
    DRDG \cite{liu2021drdg} & 31.28 & 71.50 & 33.35 & 69.14 \\
    ANRL \cite{liu2021anrl} & 31.06 & 72.12 & 30.73 & 74.10 \\
    SSAN \cite{wang2022ssan} & 30.00 & 76.20 & 29.44 & 76.62 \\
    EBDG \cite{du2022ebdg} & 27.97 & 75.84 & 25.94 & 78.28 \\
    AMEL \cite{zhou2022amel} & 24.52 & 82.12 & 19.68 & 87.01 \\
    IAGD \cite{zhou2023iadg} & 24.07 & 85.13 &18.47 &\textbf{90.49} \\
    \hline
    \rowcolor{backcolour} \SystemName~(\textit{\textbf{ours}}) &  \textbf{16.91}$^{1.17}$ & \textbf{88.12}$^{0.58}$ & \textbf{17.88}$^{0.15}$ & {89.67}$^{0.39}$ \\
    \hline \hline
    \end{tabular}%
    }
\caption{{Evaluation on limited source domains.} Baseline results are sourced from \cite{zhou2023iadg}.}
\label{tab:limited_source}

\end{table}

 \subsection{Experiment Settings}
\textbf{Datasets.} Our experiments are conducted on four benchmark datasets: \texttt{Idiap Replay Attack} \cite{chingovska2012indiap_replay} (\textbf{I}), \texttt{OULU-NPU} \cite{boulkenafet2017oulu_npu} (\textbf{O}), \texttt{CASIA-MFSD} \cite{zhang2012casia} (\textbf{C}), and \texttt{MSU-MFSD} \cite{wen2015msu_mfsd} (\textbf{M}). Consistent with prior works, we treat each dataset as a separate domain and employ a leave-one-out testing protocol to evaluate cross-domain generalization capabilities. For instance, the protocol \textbf{ICM $\rightarrow$ O} involves training on \texttt{Idiap Replay Attack}, \texttt{CASIA-MFSD}, and \texttt{MSU-MFSD}, and testing on \texttt{OULU-NPU}.
\begin{figure}[t!]
\begin{minipage}{0.25\textwidth}
\centering
\resizebox{.92\textwidth}{!}{%
\begin{tabular}{l|c}
	\hline
	Methods & AUC$\uparrow$   \\
	\hline \hline
	SVM1+IMQ \cite{arashloo2017anomaly} & 70.23$^{\text{12.69}}$  \\
	CDCN \cite{yu2020searching} & {88.69}$^{\text{10.56}}$  \\
	CDCN++ \cite{yu2020searching} & {87.53}$^{\text{10.90}}$ \\
	SSAN \cite{wang2022ssan} & {88.01}$^{\text{9.93}}$ \\
	TTN-S \cite{wang2022learning} &{89.71}$^{\text{9.17}}$ \\
	UDG-FAS \cite{liu2023udgfasssdg} & {92.43}$^{\text{6.86}}$ \\
    \hline
    \rowcolor{backcolour} GAC-FAS (\textit{\textbf{ours}}) &    \textbf{93.39}$^{\text{4.27}}$ \\
	\hline
	\hline
\end{tabular}%
}
\subcaption{Unseen 2D attack}
\end{minipage}%
\begin{minipage}{0.25\textwidth}
\centering
\resizebox{.90\textwidth}{!}{%
\begin{tabular}{l|l}
	\hline
	Method & AUC$\uparrow$   \\
	\hline \hline
	Saha \textit{et al.} \cite{saha2020domain} & 79.20 \\
    Panwar \textit{et al.} \cite{panwar2021unsupervised} & 80.00 \\
    \hline
    SSDG-R \cite{jia2020ssgd} & 82.11 \\
    CIFAS \cite{liu2022causal} & 83.20 \\
    UDG-FAS \cite{liu2023udgfasssdg} & 87.26 \\
	\hline
    \rowcolor{backcolour} GAC-FAS (\textit{\textbf{ours}}) &  \textbf{89.27}$^{\text{0.58}}$ \\
	\hline
	\hline
\end{tabular}%
}

\subcaption{Unseen 3D attack}
\end{minipage}
\captionof{table}{{Evaluation on unseen attacks.} Baseline results are sourced from \cite{liu2023udgfasssdg}.}
\addtocounter{figure}{-1}
\label{tab:unseen}
\end{figure}
\begin{table*}[t!]
\centering
\resizebox{.99\textwidth}{!}{%
    \begin{tabular}{l|l|l|l|l}
    \hline
    \multirow{2}{*}{ Methods } &  \multicolumn{1}{c |}{ \textbf{ICM {$\rightarrow$} O} } & \multicolumn{1}{c |}{  \textbf{OCM {$\rightarrow$} I} } & \multicolumn{1}{c |}{\textbf{OCI {$\rightarrow$} M} } & \multicolumn{1}{c}{ \textbf{OMI {$\rightarrow$} C} } \\ 
    & HTER $\downarrow / \mathrm{AUC} \uparrow / \mathrm{TPR} 95 \uparrow$ & HTER $\downarrow / \mathrm{AUC} \uparrow / \mathrm{TPR} 95 \uparrow$& HTER $\downarrow / \mathrm{AUC} \uparrow / \mathrm{TPR} 95 \uparrow$ & HTER $\downarrow / \mathrm{AUC} \uparrow / \mathrm{TPR} 95 \uparrow$ \\
    \hline
    \hline
     SSDG-R \cite{jia2020ssgd}   & 15.83$^{1.29}$ / 92.13$^{0.96}$ / 66.54$^{4.00}$ & 14.65$^{1.21}$ / 91.93$^{1.35}$  / 53.68$^{2.56}$ & 22.84$^{1.14}$  / 78.67$^{1.31}$  / 50.80$^{5.95}$ &  28.76$^{0.89}$ / 80.91$^{1.10}$  / 41.47$^{2.68}$\\
     SSAN-R  \cite{wang2022ssan}  & 25.72$^{3.74}$ / 79.37$^{4.69}$ / 36.75$^{5.19}$ & 35.39$^{8.04}$ / 70.13$^{9.03}$ / 64.00$^{2.70}$ &  21.79$^{3.68}$  / 84.06$^{3.78}$ / 51.91$^{4.28}$ & 26.44$^{2.91}$ / 78.84$^{2.83}$ / 45.36$^{4.29}$\\
     PatchNet \cite{wang2022patchnet}  & 23.49$^{1.80}$ / 84.62$^{1.92}$ / 39.39$^{6.83}$ & 29.75$^{2.76}$ / 80.53$^{1.35}$ / 54.25$^{2.18}$& 25.92$^{1.13}$ / 83.43$^{0.87}$ / 38.75$^{8.31}$ & 36.26$^{1.98}$ / 71.38$^{1.89}$ / 19.22$^{3.85}$\\
     SA-FAS \cite{sun2023safas} & {11.29}$^{0.32}$ / {95.23}$^{0.24}$ / {73.38}$^{1.64}$ & \textbf{11.48}$^{1.10}$ / \textbf{95.74}$^{0.55}$ / {77.05}$^{3.26}$ & {14.36}$^{1.10}$ / {92.06}$^{0.53}$ / {55.71}$^{4.82}$ & {19.40}$^{0.66}$ / {88.69}$^{0.67}$ / {50.53}$^{3.60}$ \\
    \hline
    \rowcolor{backcolour} \SystemName\ (\textit{\textbf{ours}}) & \textbf{9.89 }$^{0.47}$ / \textbf{96.44}$^{0.18}$ / \textbf{80.47}$^{1.34}$ & 12.51$^{3.03}$ / {93.03}$^{2.24}$ / \textbf{77.38}$^{8.50}$ &\textbf{12.29}$^{1.29}$ / \textbf{95.35}$^{0.57}$/ \textbf{72.00}$^{3.84}$ &   \textbf{15.37}$^{1.52}$ / \textbf{91.67}$^{1.67}$/ \textbf{58.67}$^{10.55}$ \\
    \hline
    \hline
    \end{tabular}%
    }
\caption{{Evaluation at convergence.} A comprehensive assessment of cross-domain face anti-spoofing on prominent databases: CASIA (\textbf{C}), Idiap Replay (\textbf{I}), MSU-MFSD (\textbf{M}), and Oulu-NPU (\textbf{O}). Methods are benchmarked using their mean and standard deviation performance over the final 10 evaluations. Baseline results are sourced from \cite{sun2023safas}.} 
\label{tab:convergence}

\end{table*}
\textbf{Implementation Details.} Input images are detected and cropped using MTCNN \cite{zhang2016mtcnn}, then resized to 256$\times$256 pixels. We employ a ResNet-18 \cite{he2016deep} architecture, pre-trained on the ImageNet dataset \cite{ILSVRC15}, as our feature extraction backbone to maintain consistency with SoTA baselines \cite{sun2023safas, zhou2023iadg, wang2022ssan, jia2020ssgd}. The network is trained using an SGD optimizer with an initial learning rate of 0.005. Our regularization strategy includes weight decay and supervised contrastive learning, applied at an intermediate layer, to promote inter-domain discriminability \cite{khosla2020supervised, sun2023safas}. The hyperparameters are set as follows: $\{\gamma=0.0002, \rho=0.1\}$. We run each experiment three times and take the average performance to report.

\textbf{Evaluation Metrics.} Model performance is quantified using three standard metrics: Half Total Error Rate (HTER), Area Under the Receiver Operating Characteristic Curve (AUC), and True Positive Rate (TPR) at a False Positive Rate (FPR) of 5\%, denoted as TPR95.

\subsection{Comparison to SoTA baselines}
\label{subsec:sota_compare}

 \textbf{Leave-One-Out}. 
 Table \ref{tab:main_results} presents a comprehensive comparison with a broad range of recent studies addressing DG in FAS. Based on the results, we make the following observations:
(1) Recent advancements in FAS methods have achieved significant breakthroughs in performance. However, a performance plateau is evident among these methods, largely because they do not incorporate the sharpness of the loss landscape into their objectives.
(2) Our method consistently outperforms the majority of the surveyed DG for FAS approaches \cite{liu2023udgfasssdg, sun2023safas, wang2022patchnet, zhou2023iadg, wang2022ssan, jia2020ssgd} across all four experimental setups. Notably, we report a 1.56\% improvement in the HTER (reducing from 5.86\% to 4.29\%) in the \textbf{OCM $\rightarrow$ I} experiment, translating to an enhancement of over 26\%.

\begin{table}[!t]
\centering
\resizebox{0.48\textwidth}{!}{%
\begin{tabular}{c!{\color{black}\VRule[1pt]} c !{\color{black}\VRule[1pt]} c   c c }
        \hline
        {Model / loss} & DGrad &   HTER $\downarrow$ & {AUC} $\uparrow$ & TPR95 $\uparrow$ \tabularnewline
        \hline
        \hline
         Fish \cite{shi2021gradient} & \checkmark  & {33.83}$^{0.74}$ & {72.34}$^{0.37}$ & {17.50}$^{1.48}$\Tstrut\tabularnewline 
        \hline
        \multirow{2}{*}{ SAM \cite{foret2020sharpness} } & \xmark  &{11.99}$^{0.49}$ & {95.13}$^{0.16}$ & {73.31}$^{1.57}$\Tstrut\tabularnewline
        & \checkmark & {12.51}$^{0.89}$ & {95.45}$^{0.35}$ & {72.94}$^{2.99}$\Tstrut\tabularnewline
        \hline
        \multirow{2}{*}{ ASAM \cite{kwon2021asam} } & \xmark  & {11.73}$^{0.40}$ & {95.36}$^{0.28}$ & {75.36}$^{1.61}$\Tstrut\tabularnewline
         & \checkmark & {12.18}$^{0.93}$ & {95.40}$^{0.49}$ & {75.92}$^{1.54}$\Tstrut\tabularnewline
        \hline
        \multirow{2}{*}{ SAGM \cite{wang2023sagm} }& \xmark  & {11.63}$^{0.53}$ & {95.33}$^{0.23}$ & {74.81}$^{2.07}$ \Tstrut\tabularnewline 
         & \checkmark & {12.19}$^{0.50}$ & {95.06}$^{0.21}$ & {74.17}$^{1.48}$ \Tstrut\tabularnewline 
        \hline
        \multirow{2}{*}{ LookSAM \cite{liu2022looksam}  }& \xmark & {11.56}$^{0.54}$ & {95.56}$^{0.15}$ & {75.64}$^{1.19}$\Tstrut\tabularnewline        
         & \checkmark & {11.06}$^{0.50}$ & {95.76}$^{0.15}$ & {75.86}$^{1.75}$\Tstrut\tabularnewline   
        \hline
        \multirow{2}{*}{ GSAM \cite{zhuang2021gsam} } & \xmark  & {11.60}$^{0.38}$ & {95.30}$^{0.18}$ & {74.44}$^{1.79}$ \Tstrut\tabularnewline              
         & \checkmark  & {12.48}$^{0.83}$ & {95.54}$^{0.35}$ & {73.39}$^{2.92}$ \Tstrut\tabularnewline 
        \hline   
        
        \multicolumn{1}{c|}{{\small{Reg. $\langle\nabla\mathcal{L}_{\text{p}_i}(\mathcal{S}_i), \nabla\mathcal{L}(\boldsymbol{\mathcal{S}})\rangle$}}}  & \checkmark & {11.35}$^{0.54}$ & {95.55}$^{0.18}$ & {73.58}$^{1.21}$ \Tstrut\tabularnewline 
        
        \rowcolor{backcolour}\multicolumn{1}{c|}{\SystemName~\textit{(\textbf{ours})}}  & \checkmark  & \textbf{9.89 }$^{0.47}$ & \textbf{96.44}$^{0.18}$ & \textbf{80.47}$^{1.34}$ \Tstrut\tabularnewline
        \hline
        \end{tabular}%
}
        \caption{\textbf{Ablation study:}  Alternative approach using domain gradient  for DG in FAS. DGrad indicates whether domain-wise gradients are used in each method (see Fig. \ref{fig:objective}).} 
        \label{tab:ablation_loss}
\end{table}

\textbf{Limited Source Domains}.
Table \ref{tab:limited_source} summarizes the performance of our method when source domains are extremely limited. Adhering to the standard settings established by prior research \cite{zhou2023iadg, zhou2022amel, wang2022ssan}, we utilize \texttt{MSU-MFSD} and \texttt{Idiap Replay Attack} as source domains for training, while \texttt{OULU-NPU} and \texttt{CASIA-MFSD} serve as the {test datasets}. Our proposed method consistently surpasses  SoTA  baselines, achieving substantial margins of improvement on the HTER metric. A particularly significant enhancement is observed on the \textbf{MI $\rightarrow$ C} setup, where our approach yields approximately a 7\% reduction in HTER (decreasing from 24.07\% to 16.91\%).
\begin{figure}[t!]
\centering
\includegraphics[width=8cm]{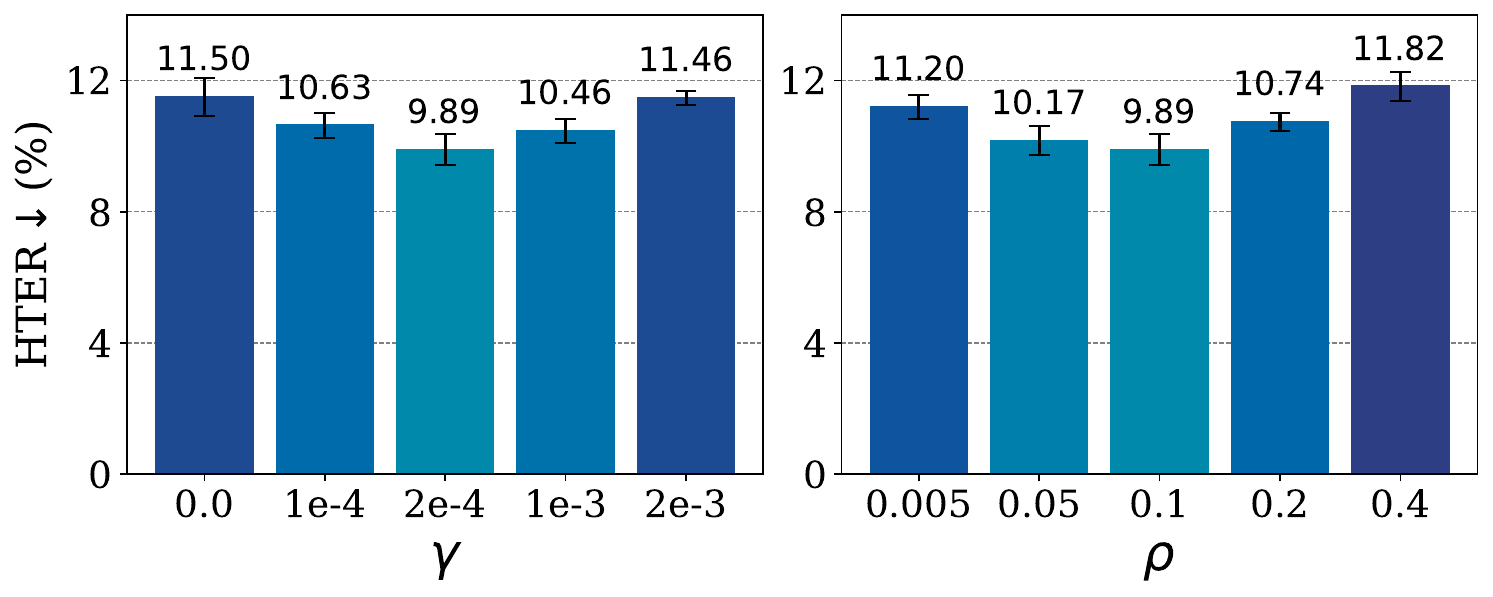}
\caption{\textbf{Ablation study: }Sensitivity analysis of hyper-parameters $\gamma$ and $\rho$ on \textbf{ICM $\rightarrow$ O} upon convergence performance.}
\label{fig:ablation_alpha_rho}

\end{figure}

\begin{figure*}[t!]
\centering
\includegraphics[width=17.3cm]{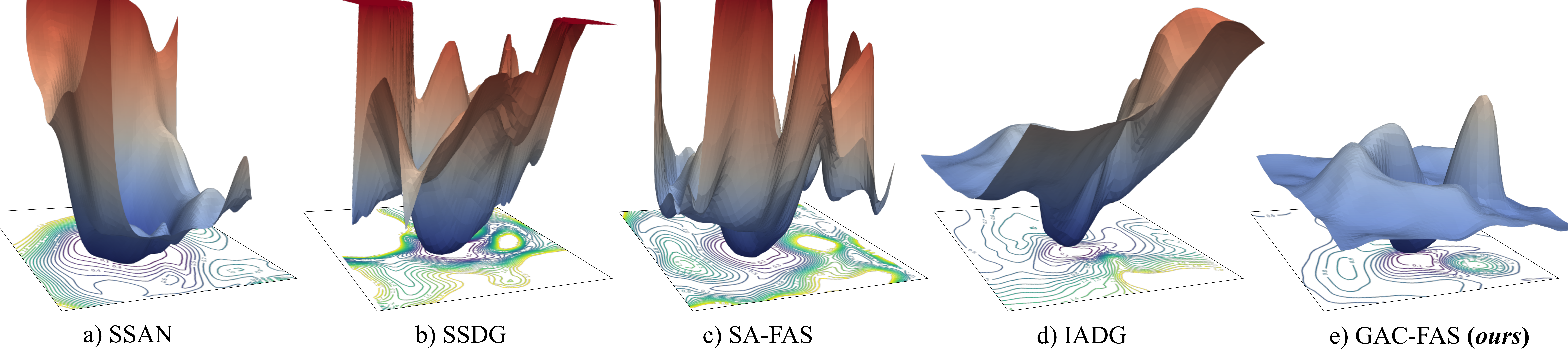}
\caption{\textbf{Loss Landscape Visualization}~\cite{li2018visualizing}. SoTA baselines exhibit convergence to sharp minima (Figures \textbf{a} to \textbf{d}) on the training set of ICM$\rightarrow$O. In contrast, our \SystemName\ (Figure \textbf{e}) achieves convergence to a flatter minimum, indicative of potentially better generalization.}
\label{fig:loss_landscape}
\end{figure*}
\textbf{Unseen Attacks}. In this experiment, we evaluate the detector's performance against unseen 2D and 3D attacks. For 2D attacks, we adopt the `leave-one-attack-type-out' method from \cite{arashloo2017anomaly}, training on two domains of  I, C, M, and testing on an unseen attack of unseen domain. In 3D attacks, we train on O, C, M, and evaluate using a 3D attack subset from the CelebA-Spoof dataset\cite{zhang2020celeba}. Results, shown in Table \ref{tab:unseen}, indicate our approach outperforms baselines by significant margins of 0.96\% and 2.01\% on AUC metric for 2D and 3D attacks, respectively.

\textbf{Comparison Upon Convergence }.
In their recent study, Sun \textit{et al.} highlight that a snapshot performance report of a test set may not accurately reflect the true generalization ability of a detection model \cite{sun2023safas}. In alignment with their methodology, we report the average performance of our model across the last 10 evaluations in Table \ref{tab:convergence}. As evident from the results, our method consistently demonstrates superior convergence on the three metrics: HTER, AUC, and TPR95, and achieves comparable results to SA-FAS in the OCM $\rightarrow$ I experiment. Notably, we observe a 4\% reduction in HTER in the OMI $\rightarrow$ C experiment (from 19.40\% down to 15.37\%). These results suggest that our proposed \SystemName\ enables the model to converge to flatter and more stable minima compared to other approaches.

\subsection{Ablation Studies}
\textbf{Alternatives of Optimizer.} In Table \ref{tab:ablation_loss}, we explore various alternative objectives for DG in FAS, including SAM \cite{foret2020sharpness}, ASAM \cite{kwon2021asam}, GSAM \cite{zhuang2021gsam}, \cite{wang2023sagm}, and LookSAM \cite{liu2022looksam}. Furthermore, we incorporate Fish \cite{shi2021gradient}, a gradient matching method for DG, into our study. We experiment with ICM$\rightarrow$ O task and report their performances upon convergence. The checkmarks in the second column denote the utilization of domain-wise gradients (see Fig. \ref{fig:objective}). While most methods demonstrate improved convergence with whole-data gradients per iteration compared to domain-wise gradients, they do not unequivocally surpass the existing SoTA baselines in DG for FAS. Domain-wise gradients, as we observed, are prone to noisy gradients at ascending points, whereas whole-data gradients tend to be dominated by specific domains, which can impede convergence. Notably, Fish \cite{shi2021gradient} exhibits competitive performance on other datasets but shows slower convergence on FAS datasets compared to SAM-based objectives.

\textbf{{Effects} of Hyper-parameters $\gamma$ and $\rho$}.
We investigate the sensitivities of our \SystemName\ with respect to $\gamma$ and $\rho$, summarizing the results of our analysis in Fig. \ref{fig:ablation_alpha_rho}. This study focuses on the \textbf{ICM $\rightarrow$ O} task, reporting the HTER metric performance upon convergence. We test a range of values for the hyperparameters $\gamma \in \{0.0, 0.0001, 0.0002, 0.001, 0.002\}$ and $\rho \in \{0.005, 0.05, 0.1, 0.2, 0.4\}$, noting that at $\gamma = 0.0$, no regularization is applied. The results indicate that with $\gamma$ set to 0.0, our model's performance is competitive with the current SoTA as shown in Table \ref{tab:convergence}, yet it does not achieve the best result. Furthermore, when the settings for $\gamma$ and $\rho$ are not excessively large, the model's performance tends to be stable and can exceed that of previous SoTA methods. However, higher values of these hyperparameters may deteriorate performance. It is important to note that we did not fine-tune these hyperparameters to optimize test accuracy for each experimental task; thus, the same hyper-parameter settings used in Sec. \ref{subsec:sota_compare} may not be optimal, even though they outperform all current SoTA methods on the datasets.

\textbf{Loss Landscape Visualization}.
Figure~\ref{fig:loss_landscape} presents the loss landscape visualization~\cite{li2018visualizing} of \SystemName, in comparison with four SoTA approaches: SSAN~\cite{wang2022ssan}, SSDG~\cite{liu2023udgfasssdg}, SA-FAS~\cite{sun2023safas}, and IADG~\cite{zhou2023iadg}.  We employ negative log-likelihood as the loss metric and use the training set of the ICM$\rightarrow$O task for visualization. While all baseline methods demonstrate comparable generalization capabilities to our method in Table~\ref{tab:main_results}, they distinctly exhibit sharp minima in their loss functions, characterized by steep gradients in the loss landscape as shown in Fig.~\ref{fig:loss_landscape}.\ \textbf{a)}-\textbf{d)}. In contrast, our proposed approach reveals a flatter minimum, which may correlate with enhanced generalization. These observations provide additional insights into the superior numerical results achieved by our method in various experiments, as discussed in Sec.~\ref{subsec:sota_compare}.

\section{Conclusion}
\label{sec:conclusion}
In this paper, we introduced \SystemName, a novel framework designed to optimize the minimum for Domain Generalization (DG) in Face Anti-Spoofing (FAS). Inspired by recent advancements in DG methods that leverage loss sharpness-aware objectives, our approach involves identifying and utilizing ascending points for each domain within the training dataset. {A key underlying novelty} in our methodology is the regulation of SAM generalization gradients of whole data at these ascending points, aligning them coherently with gradients derived from ERM.
Through comprehensive analysis and a series of rigorous experiments, we have demonstrated that \SystemName\ not only achieves superior generalization capabilities in FAS tasks but also consistently outperforms current SoTA baselines by significant margins. This performance consistency is observed across a variety of common experimental setups, underscoring the robustness and effectiveness of our proposed method. Our limitations and further analyses are provided in the Supp. material to support future studies.

{\small
\bibliographystyle{ieee_fullname}
\bibliography{egbib}

\begin{thebibliography}{10}\itemsep=-1pt

\bibitem{arashloo2017anomaly}
Shervin~Rahimzadeh Arashloo, Josef Kittler, and William Christmas.
\newblock An anomaly detection approach to face spoofing detection: A new formulation and evaluation protocol.
\newblock {\em IEEE access}, 5:13868--13882, 2017.

\bibitem{boulkenafet2015face}
Zinelabidine Boulkenafet, Jukka Komulainen, and Abdenour Hadid.
\newblock Face anti-spoofing based on color texture analysis.
\newblock In {\em 2015 IEEE international conference on image processing (ICIP)}, pages 2636--2640. IEEE, 2015.

\bibitem{boulkenafet2017oulu_npu}
Zinelabinde Boulkenafet, Jukka Komulainen, Lei Li, Xiaoyi Feng, and Abdenour Hadid.
\newblock Oulu-npu: A mobile face presentation attack database with real-world variations.
\newblock In {\em 2017 12th IEEE international conference on automatic face \& gesture recognition (FG 2017)}, pages 612--618. IEEE, 2017.

\bibitem{cha2021swad}
Junbum Cha, Sanghyuk Chun, Kyungjae Lee, Han-Cheol Cho, Seunghyun Park, Yunsung Lee, and Sungrae Park.
\newblock Swad: Domain generalization by seeking flat minima.
\newblock {\em Advances in Neural Information Processing Systems}, 34:22405--22418, 2021.

\bibitem{chen2021d2am}
Zhihong Chen, Taiping Yao, Kekai Sheng, Shouhong Ding, Ying Tai, Jilin Li, Feiyue Huang, and Xinyu Jin.
\newblock Generalizable representation learning for mixture domain face anti-spoofing.
\newblock In {\em Proceedings of the AAAI conference on artificial intelligence}, volume~35, pages 1132--1139, 2021.

\bibitem{chetty2010biometric}
Girija Chetty.
\newblock Biometric liveness checking using multimodal fuzzy fusion.
\newblock In {\em International Conference on Fuzzy Systems}, pages 1--8. IEEE, 2010.

\bibitem{chingovska2012indiap_replay}
Ivana Chingovska, Andr{\'e} Anjos, and S{\'e}bastien Marcel.
\newblock On the effectiveness of local binary patterns in face anti-spoofing.
\newblock In {\em 2012 BIOSIG-proceedings of the international conference of biometrics special interest group (BIOSIG)}, pages 1--7. IEEE, 2012.

\bibitem{de2013lbp}
Tiago de Freitas~Pereira, Andr{\'e} Anjos, Jos{\'e}~Mario De~Martino, and S{\'e}bastien Marcel.
\newblock Lbp- top based countermeasure against face spoofing attacks.
\newblock In {\em Computer Vision-ACCV 2012 Workshops: ACCV 2012 International Workshops, Daejeon, Korea, November 5-6, 2012, Revised Selected Papers, Part I 11}, pages 121--132. Springer, 2013.

\bibitem{deng2019arcface}
Jiankang Deng, Jia Guo, Niannan Xue, and Stefanos Zafeiriou.
\newblock Arcface: Additive angular margin loss for deep face recognition.
\newblock In {\em Proceedings of the IEEE/CVF Conference on Computer Vision and Pattern Recognition}, pages 4690--4699, 2019.

\bibitem{du2021esam}
Jiawei Du, Hanshu Yan, Jiashi Feng, Joey~Tianyi Zhou, Liangli Zhen, Rick Siow~Mong Goh, and Vincent Tan.
\newblock Efficient sharpness-aware minimization for improved training of neural networks.
\newblock In {\em International Conference on Learning Representations}, 2021.

\bibitem{du2022ebdg}
Zhekai Du, Jingjing Li, Lin Zuo, Lei Zhu, and Ke Lu.
\newblock Energy-based domain generalization for face anti-spoofing.
\newblock In {\em Proceedings of the 30th ACM International Conference on Multimedia}, pages 1749--1757, 2022.

\bibitem{dziugaite2017computing}
Gintare~Karolina Dziugaite and Daniel~M Roy.
\newblock Computing nonvacuous generalization bounds for deep (stochastic) neural networks with many more parameters than training data.
\newblock {\em Uncertainty in Artificial Intelligence}, 2017.

\bibitem{feng2016integration}
Litong Feng, Lai-Man Po, Yuming Li, Xuyuan Xu, Fang Yuan, Terence Chun-Ho Cheung, and Kwok-Wai Cheung.
\newblock Integration of image quality and motion cues for face anti-spoofing: A neural network approach.
\newblock {\em Journal of Visual Communication and Image Representation}, 38:451--460, 2016.

\bibitem{foret2020sharpness}
Pierre Foret, Ariel Kleiner, Hossein Mobahi, and Behnam Neyshabur.
\newblock Sharpness-aware minimization for efficiently improving generalization.
\newblock In {\em International Conference on Learning Representations}, 2020.

\bibitem{garipov2018loss}
Timur Garipov, Pavel Izmailov, Dmitrii Podoprikhin, Dmitry~P Vetrov, and Andrew~G Wilson.
\newblock Loss surfaces, mode connectivity, and fast ensembling of dnns.
\newblock {\em Advances in neural information processing systems}, 31, 2018.

\bibitem{george2019deep}
Anjith George and S{\'e}bastien Marcel.
\newblock Deep pixel-wise binary supervision for face presentation attack detection.
\newblock In {\em 2019 International Conference on Biometrics (ICB)}, pages 1--8. IEEE, 2019.

\bibitem{fool3d}
Andy Greenberg.
\newblock Hackers say they've broken face id a week after iphone x release.
\newblock \url{https://www.wired.com/story/hackers-say-broke-face-id-security/}, November 2017.
\newblock Accessed: 2023-07-01.

\bibitem{guo2022multi}
Xiao Guo, Yaojie Liu, Anil Jain, and Xiaoming Liu.
\newblock Multi-domain learning for updating face anti-spoofing models.
\newblock In {\em European Conference on Computer Vision}, pages 230--249. Springer, 2022.

\bibitem{he2016deep}
Kaiming He, Xiangyu Zhang, Shaoqing Ren, and Jian Sun.
\newblock Deep residual learning for image recognition.
\newblock In {\em Proceedings of the IEEE conference on computer vision and pattern recognition}, pages 770--778, 2016.

\bibitem{hochreiter1994simplifying}
Sepp Hochreiter and J{\"u}rgen Schmidhuber.
\newblock Simplifying neural nets by discovering flat minima.
\newblock {\em Advances in neural information processing systems}, 7, 1994.

\bibitem{huang2022adaptive}
Hsin-Ping Huang, Deqing Sun, Yaojie Liu, Wen-Sheng Chu, Taihong Xiao, Jinwei Yuan, Hartwig Adam, and Ming-Hsuan Yang.
\newblock Adaptive transformers for robust few-shot cross-domain face anti-spoofing.
\newblock In {\em European Conference on Computer Vision}, pages 37--54. Springer, 2022.

\bibitem{izmailov2018averaging}
Pavel Izmailov, Dmitrii Podoprikhin, Timur Garipov, Dmitry Vetrov, and Andrew~Gordon Wilson.
\newblock Averaging weights leads to wider optima and better generalization.
\newblock In {\em 34th Conference on Uncertainty in Artificial Intelligence 2018, UAI 2018}, pages 876--885. Association For Uncertainty in Artificial Intelligence (AUAI), 2018.

\bibitem{jia2020ssgd}
Yunpei Jia, Jie Zhang, Shiguang Shan, and Xilin Chen.
\newblock Single-side domain generalization for face anti-spoofing.
\newblock In {\em Proceedings of the IEEE/CVF Conference on Computer Vision and Pattern Recognition}, pages 8484--8493, 2020.

\bibitem{jiang2020deeperforensics}
Liming Jiang, Ren Li, Wayne Wu, Chen Qian, and Chen~Change Loy.
\newblock Deeperforensics-1.0: A large-scale dataset for real-world face forgery detection.
\newblock In {\em Proceedings of the IEEE/CVF conference on computer vision and pattern recognition}, pages 2889--2898, 2020.

\bibitem{keskar2016large}
Nitish~Shirish Keskar, Dheevatsa Mudigere, Jorge Nocedal, Mikhail Smelyanskiy, and Ping Tak~Peter Tang.
\newblock On large-batch training for deep learning: Generalization gap and sharp minima.
\newblock In {\em International Conference on Learning Representations}, 2016.

\bibitem{khosla2020supervised}
Prannay Khosla, Piotr Teterwak, Chen Wang, Aaron Sarna, Yonglong Tian, Phillip Isola, Aaron Maschinot, Ce Liu, and Dilip Krishnan.
\newblock Supervised contrastive learning.
\newblock {\em Advances in neural information processing systems}, 33:18661--18673, 2020.

\bibitem{kingma2014adam}
Diederik~P Kingma and Jimmy Ba.
\newblock Adam: A method for stochastic optimization.
\newblock {\em arXiv preprint arXiv:1412.6980}, 2014.

\bibitem{kollreider2007real}
Klaus Kollreider, Hartwig Fronthaler, Maycel~Isaac Faraj, and Josef Bigun.
\newblock Real-time face detection and motion analysis with application in “liveness” assessment.
\newblock {\em IEEE Transactions on Information Forensics and Security}, 2(3):548--558, 2007.

\bibitem{komulainen2013context}
Jukka Komulainen, Abdenour Hadid, and Matti Pietik{\"a}inen.
\newblock Context based face anti-spoofing.
\newblock In {\em 2013 IEEE Sixth International Conference on Biometrics: Theory, Applications and Systems (BTAS)}, pages 1--8. IEEE, 2013.

\bibitem{fool2d}
Paul Kunert.
\newblock Phones' facial recog tech `fooled' by low-res 2d photo.
\newblock \url{https://www.theregister.com/2023/05/19/2d_photograph_facial_recog/}, May 2023.
\newblock Accessed: 2023-07-01.

\bibitem{kwon2021asam}
Jungmin Kwon, Jeongseop Kim, Hyunseo Park, and In~Kwon Choi.
\newblock Asam: Adaptive sharpness-aware minimization for scale-invariant learning of deep neural networks.
\newblock In {\em International Conference on Machine Learning}, pages 5905--5914. PMLR, 2021.

\bibitem{le2023quality}
Binh~M Le and Simon~S Woo.
\newblock Quality-agnostic deepfake detection with intra-model collaborative learning.
\newblock In {\em Proceedings of the IEEE/CVF International Conference on Computer Vision}, pages 22378--22389, 2023.

\bibitem{li2018unsupervised}
Haoliang Li, Wen Li, Hong Cao, Shiqi Wang, Feiyue Huang, and Alex~C Kot.
\newblock Unsupervised domain adaptation for face anti-spoofing.
\newblock {\em IEEE Transactions on Information Forensics and Security}, 13(7):1794--1809, 2018.

\bibitem{li2018mmdaae}
Haoliang Li, Sinno~Jialin Pan, Shiqi Wang, and Alex~C Kot.
\newblock Domain generalization with adversarial feature learning.
\newblock In {\em Proceedings of the IEEE conference on computer vision and pattern recognition}, pages 5400--5409, 2018.

\bibitem{li2018visualizing}
Hao Li, Zheng Xu, Gavin Taylor, Christoph Studer, and Tom Goldstein.
\newblock Visualizing the loss landscape of neural nets.
\newblock {\em Advances in neural information processing systems}, 31, 2018.

\bibitem{li2016original}
Lei Li, Xiaoyi Feng, Zinelabidine Boulkenafet, Zhaoqiang Xia, Mingming Li, and Abdenour Hadid.
\newblock An original face anti-spoofing approach using partial convolutional neural network.
\newblock In {\em 2016 Sixth international conference on image processing theory, tools and applications (IPTA)}, pages 1--6. IEEE, 2016.

\bibitem{liao2023domain}
Chen-Hao Liao, Wen-Cheng Chen, Hsuan-Tung Liu, Yi-Ren Yeh, Min-Chun Hu, and Chu-Song Chen.
\newblock Domain invariant vision transformer learning for face anti-spoofing.
\newblock In {\em Proceedings of the IEEE/CVF Winter Conference on Applications of Computer Vision}, pages 6098--6107, 2023.

\bibitem{lin2019face}
Bofan Lin, Xiaobai Li, Zitong Yu, and Guoying Zhao.
\newblock Face liveness detection by rppg features and contextual patch-based cnn.
\newblock In {\em Proceedings of the 2019 3rd international conference on biometric engineering and applications}, pages 61--68, 2019.

\bibitem{liu2021taming}
Haozhe Liu, Zhe Kong, Raghavendra Ramachandra, Feng Liu, Linlin Shen, and Christoph Busch.
\newblock Taming self-supervised learning for presentation attack detection: In-image de-folding and out-of-image de-mixing.
\newblock {\em arXiv preprint arXiv:2109.04100}, 2021.

\bibitem{liu2021anrl}
Shubao Liu, Ke-Yue Zhang, Taiping Yao, Mingwei Bi, Shouhong Ding, Jilin Li, Feiyue Huang, and Lizhuang Ma.
\newblock Adaptive normalized representation learning for generalizable face anti-spoofing.
\newblock In {\em Proceedings of the 29th ACM international conference on multimedia}, pages 1469--1477, 2021.

\bibitem{liu2021drdg}
Shubao Liu, Ke-Yue Zhang, Taiping Yao, Kekai Sheng, Shouhong Ding, Ying Tai, Jilin Li, Yuan Xie, and Lizhuang Ma.
\newblock Dual reweighting domain generalization for face presentation attack detection.
\newblock In {\em Proceedings of the Thirtieth International Joint Conference on Artificial Intelligence}, pages 867--873, 2021.

\bibitem{liu2017sphereface}
Weiyang Liu, Yandong Wen, Zhiding Yu, Ming Li, Bhiksha Raj, and Le Song.
\newblock Sphereface: Deep hypersphere embedding for face recognition.
\newblock In {\em Proceedings of the IEEE conference on computer vision and pattern recognition}, pages 212--220, 2017.

\bibitem{liu2022causal}
Yuchen Liu, Yabo Chen, Wenrui Dai, Chenglin Li, Junni Zou, and Hongkai Xiong.
\newblock Causal intervention for generalizable face anti-spoofing.
\newblock In {\em 2022 IEEE International Conference on Multimedia and Expo (ICME)}, pages 01--06. IEEE, 2022.

\bibitem{liu2023udgfasssdg}
Yuchen Liu, Yabo Chen, Mengran Gou, Chun-Ting Huang, Yaoming Wang, Wenrui Dai, and Hongkai Xiong.
\newblock Towards unsupervised domain generalization for face anti-spoofing.
\newblock In {\em Proceedings of the IEEE/CVF International Conference on Computer Vision}, pages 20654--20664, 2023.

\bibitem{liu2023towards}
Yuchen Liu, Yabo Chen, Mengran Gou, Chun-Ting Huang, Yaoming Wang, Wenrui Dai, and Hongkai Xiong.
\newblock Towards unsupervised domain generalization for face anti-spoofing.
\newblock In {\em Proceedings of the IEEE/CVF International Conference on Computer Vision}, pages 20654--20664, 2023.

\bibitem{liu2022looksam}
Yong Liu, Siqi Mai, Xiangning Chen, Cho-Jui Hsieh, and Yang You.
\newblock Towards efficient and scalable sharpness-aware minimization.
\newblock In {\em Proceedings of the IEEE/CVF Conference on Computer Vision and Pattern Recognition}, pages 12360--12370, 2022.

\bibitem{maatta2011mslpb}
Jukka M{\"a}{\"a}tt{\"a}, Abdenour Hadid, and Matti Pietik{\"a}inen.
\newblock Face spoofing detection from single images using micro-texture analysis.
\newblock In {\em 2011 international joint conference on Biometrics (IJCB)}, pages 1--7. IEEE, 2011.

\bibitem{pan2007eyeblink}
Gang Pan, Lin Sun, Zhaohui Wu, and Shihong Lao.
\newblock Eyeblink-based anti-spoofing in face recognition from a generic webcamera.
\newblock In {\em 2007 IEEE 11th international conference on computer vision}, pages 1--8. IEEE, 2007.

\bibitem{panwar2021unsupervised}
Ankush Panwar, Pratyush Singh, Suman Saha, Danda~Pani Paudel, and Luc Van~Gool.
\newblock Unsupervised compound domain adaptation for face anti-spoofing.
\newblock In {\em 2021 16th IEEE International Conference on Automatic Face and Gesture Recognition (FG 2021)}, pages 1--8. IEEE, 2021.

\bibitem{patel2016cross}
Keyurkumar Patel, Hu Han, and Anil~K Jain.
\newblock Cross-database face antispoofing with robust feature representation.
\newblock In {\em Biometric Recognition: 11th Chinese Conference, CCBR 2016, Chengdu, China, October 14-16, 2016, Proceedings 11}, pages 611--619. Springer, 2016.

\bibitem{patel2016secure}
Keyurkumar Patel, Hu Han, and Anil~K Jain.
\newblock Secure face unlock: Spoof detection on smartphones.
\newblock {\em IEEE transactions on information forensics and security}, 11(10):2268--2283, 2016.

\bibitem{ILSVRC15}
Olga Russakovsky, Jia Deng, Hao Su, Jonathan Krause, Sanjeev Satheesh, Sean Ma, Zhiheng Huang, Andrej Karpathy, Aditya Khosla, Michael Bernstein, Alexander~C. Berg, and Li Fei-Fei.
\newblock {ImageNet Large Scale Visual Recognition Challenge}.
\newblock {\em International Journal of Computer Vision (IJCV)}, 115(3):211--252, 2015.

\bibitem{saha2020domain}
Suman Saha, Wenhao Xu, Menelaos Kanakis, Stamatios Georgoulis, Yuhua Chen, Danda~Pani Paudel, and Luc Van~Gool.
\newblock Domain agnostic feature learning for image and video based face anti-spoofing.
\newblock In {\em Proceedings of the IEEE/CVF Conference on Computer Vision and Pattern Recognition Workshops}, pages 802--803, 2020.

\bibitem{shao2019maddg}
Rui Shao, Xiangyuan Lan, Jiawei Li, and Pong~C Yuen.
\newblock Multi-adversarial discriminative deep domain generalization for face presentation attack detection.
\newblock In {\em Proceedings of the IEEE/CVF conference on computer vision and pattern recognition}, pages 10023--10031, 2019.

\bibitem{shao2020rfm}
Rui Shao, Xiangyuan Lan, and Pong~C Yuen.
\newblock Regularized fine-grained meta face anti-spoofing.
\newblock In {\em Proceedings of the AAAI Conference on Artificial Intelligence}, volume~34, pages 11974--11981, 2020.

\bibitem{shi2021gradient}
Yuge Shi, Jeffrey Seely, Philip Torr, N Siddharth, Awni Hannun, Nicolas Usunier, and Gabriel Synnaeve.
\newblock Gradient matching for domain generalization.
\newblock In {\em International Conference on Learning Representations}, 2021.

\bibitem{sun2023safas}
Yiyou Sun, Yaojie Liu, Xiaoming Liu, Yixuan Li, and Wen-Sheng Chu.
\newblock Rethinking domain generalization for face anti-spoofing: Separability and alignment.
\newblock In {\em Proceedings of the IEEE/CVF Conference on Computer Vision and Pattern Recognition}, pages 24563--24574, 2023.

\bibitem{vapnik1991erm}
Vladimir Vapnik.
\newblock Principles of risk minimization for learning theory.
\newblock {\em Advances in neural information processing systems}, 4, 1991.

\bibitem{wang2022patchnet}
Chien-Yi Wang, Yu-Ding Lu, Shang-Ta Yang, and Shang-Hong Lai.
\newblock Patchnet: A simple face anti-spoofing framework via fine-grained patch recognition.
\newblock In {\em Proceedings of the IEEE/CVF Conference on Computer Vision and Pattern Recognition}, pages 20281--20290, 2022.

\bibitem{wang2019improving}
Guoqing Wang, Hu Han, Shiguang Shan, and Xilin Chen.
\newblock Improving cross-database face presentation attack detection via adversarial domain adaptation.
\newblock In {\em 2019 International Conference on Biometrics (ICB)}, pages 1--8. IEEE, 2019.

\bibitem{wang2020unsupervised}
Guoqing Wang, Hu Han, Shiguang Shan, and Xilin Chen.
\newblock Unsupervised adversarial domain adaptation for cross-domain face presentation attack detection.
\newblock {\em IEEE Transactions on Information Forensics and Security}, 16:56--69, 2020.

\bibitem{wang2018cosface}
Hao Wang, Yitong Wang, Zheng Zhou, Xing Ji, Dihong Gong, Jingchao Zhou, Zhifeng Li, and Wei Liu.
\newblock Cosface: Large margin cosine loss for deep face recognition.
\newblock In {\em Proceedings of the IEEE conference on computer vision and pattern recognition}, pages 5265--5274, 2018.

\bibitem{wang2021sda}
Jingjing Wang, Jingyi Zhang, Ying Bian, Youyi Cai, Chunmao Wang, and Shiliang Pu.
\newblock Self-domain adaptation for face anti-spoofing.
\newblock In {\em Proceedings of the AAAI conference on artificial intelligence}, volume~35, pages 2746--2754, 2021.

\bibitem{wang2023sagm}
Pengfei Wang, Zhaoxiang Zhang, Zhen Lei, and Lei Zhang.
\newblock Sharpness-aware gradient matching for domain generalization.
\newblock In {\em Proceedings of the IEEE/CVF Conference on Computer Vision and Pattern Recognition}, pages 3769--3778, 2023.

\bibitem{wang2022learning}
Zhuo Wang, Qiangchang Wang, Weihong Deng, and Guodong Guo.
\newblock Learning multi-granularity temporal characteristics for face anti-spoofing.
\newblock {\em IEEE Transactions on Information Forensics and Security}, 17:1254--1269, 2022.

\bibitem{wang2022ssan}
Zhuo Wang, Zezheng Wang, Zitong Yu, Weihong Deng, Jiahong Li, Tingting Gao, and Zhongyuan Wang.
\newblock Domain generalization via shuffled style assembly for face anti-spoofing.
\newblock In {\em Proceedings of the IEEE/CVF Conference on Computer Vision and Pattern Recognition}, pages 4123--4133, 2022.

\bibitem{wen2015msu_mfsd}
Di Wen, Hu Han, and Anil~K Jain.
\newblock Face spoof detection with image distortion analysis.
\newblock {\em IEEE Transactions on Information Forensics and Security}, 10(4):746--761, 2015.

\bibitem{wu2020adversarial}
Dongxian Wu, Shu-Tao Xia, and Yisen Wang.
\newblock Adversarial weight perturbation helps robust generalization.
\newblock {\em Advances in Neural Information Processing Systems}, 33:2958--2969, 2020.

\bibitem{yang2013face}
Jianwei Yang, Zhen Lei, Shengcai Liao, and Stan~Z Li.
\newblock Face liveness detection with component dependent descriptor.
\newblock In {\em 2013 International Conference on Biometrics (ICB)}, pages 1--6. IEEE, 2013.

\bibitem{yu2021revisiting}
Zitong Yu, Xiaobai Li, Jingang Shi, Zhaoqiang Xia, and Guoying Zhao.
\newblock Revisiting pixel-wise supervision for face anti-spoofing.
\newblock {\em IEEE Transactions on Biometrics, Behavior, and Identity Science}, 3(3):285--295, 2021.

\bibitem{yu2020searching}
Zitong Yu, Chenxu Zhao, Zezheng Wang, Yunxiao Qin, Zhuo Su, Xiaobai Li, Feng Zhou, and Guoying Zhao.
\newblock Searching central difference convolutional networks for face anti-spoofing.
\newblock In {\em Proceedings of the IEEE/CVF Conference on Computer Vision and Pattern Recognition}, pages 5295--5305, 2020.

\bibitem{zhang2016mtcnn}
Kaipeng Zhang, Zhanpeng Zhang, Zhifeng Li, and Yu Qiao.
\newblock Joint face detection and alignment using multitask cascaded convolutional networks.
\newblock {\em IEEE signal processing letters}, 23(10):1499--1503, 2016.

\bibitem{zhang2017detecting}
Kaipeng Zhang, Zhanpeng Zhang, Hao Wang, Zhifeng Li, Yu Qiao, and Wei Liu.
\newblock Detecting faces using inside cascaded contextual cnn.
\newblock In {\em Proceedings of the IEEE International Conference on Computer Vision}, pages 3171--3179, 2017.

\bibitem{zhang2021structure}
Ke-Yue Zhang, Taiping Yao, Jian Zhang, Shice Liu, Bangjie Yin, Shouhong Ding, and Jilin Li.
\newblock Structure destruction and content combination for face anti-spoofing.
\newblock In {\em 2021 IEEE International Joint Conference on Biometrics (IJCB)}, pages 1--6. IEEE, 2021.

\bibitem{zhang2020celeba}
Yuanhan Zhang, ZhenFei Yin, Yidong Li, Guojun Yin, Junjie Yan, Jing Shao, and Ziwei Liu.
\newblock Celeba-spoof: Large-scale face anti-spoofing dataset with rich annotations.
\newblock In {\em Computer Vision--ECCV 2020: 16th European Conference, Glasgow, UK, August 23--28, 2020, Proceedings, Part XII 16}, pages 70--85. Springer, 2020.

\bibitem{zhang2012casia}
Zhiwei Zhang, Junjie Yan, Sifei Liu, Zhen Lei, Dong Yi, and Stan~Z Li.
\newblock A face antispoofing database with diverse attacks.
\newblock In {\em 2012 5th IAPR international conference on Biometrics (ICB)}, pages 26--31. IEEE, 2012.

\bibitem{zhou2023iadg}
Qianyu Zhou, Ke-Yue Zhang, Taiping Yao, Xuequan Lu, Ran Yi, Shouhong Ding, and Lizhuang Ma.
\newblock Instance-aware domain generalization for face anti-spoofing.
\newblock In {\em Proceedings of the IEEE/CVF Conference on Computer Vision and Pattern Recognition}, pages 20453--20463, 2023.

\bibitem{zhou2022amel}
Qianyu Zhou, Ke-Yue Zhang, Taiping Yao, Ran Yi, Shouhong Ding, and Lizhuang Ma.
\newblock Adaptive mixture of experts learning for generalizable face anti-spoofing.
\newblock In {\em Proceedings of the 30th ACM International Conference on Multimedia}, pages 6009--6018, 2022.

\bibitem{zhou2022generative}
Qianyu Zhou, Ke-Yue Zhang, Taiping Yao, Ran Yi, Kekai Sheng, Shouhong Ding, and Lizhuang Ma.
\newblock Generative domain adaptation for face anti-spoofing.
\newblock In {\em European Conference on Computer Vision}, pages 335--356. Springer, 2022.

\bibitem{zhou2023imbsam}
Yixuan Zhou, Yi Qu, Xing Xu, and Hengtao Shen.
\newblock Imbsam: A closer look at sharpness-aware minimization in class-imbalanced recognition.
\newblock In {\em Proceedings of the IEEE/CVF International Conference on Computer Vision}, pages 11345--11355, 2023.

\bibitem{zhou2023class}
Zhipeng Zhou, Lanqing Li, Peilin Zhao, Pheng-Ann Heng, and Wei Gong.
\newblock Class-conditional sharpness-aware minimization for deep long-tailed recognition.
\newblock In {\em Proceedings of the IEEE/CVF Conference on Computer Vision and Pattern Recognition}, pages 3499--3509, 2023.

\bibitem{zhuang2021gsam}
Juntang Zhuang, Boqing Gong, Liangzhe Yuan, Yin Cui, Hartwig Adam, Nicha~C Dvornek, James s Duncan, Ting Liu, et~al.
\newblock Surrogate gap minimization improves sharpness-aware training.
\newblock In {\em International Conference on Learning Representations}, 2021.

\end{thebibliography}
}

\begin{appendices}
\clearpage
\nocitesec{*}
In this supplementary material, we first provide a brief description of the datasets used in our experiment Section (Section \ref{sec:supp_data}). Next, the proof of Theorem 1 is provided in Section \ref{supp:prooftheorem1}. In Section \ref{sec:supp_expe},  we will detail the implementation and present ablation studies  about the detectors' robustness towards unseen corruptions.  Lastly, in Section \ref{sec:supp_limit}, we discuss the limitations of our proposed method and outline our planned future work.

\section{Datasets}
\label{sec:supp_data}
We describe here four popular benchmark datasets used to evaluate our proposed \SystemName:

\begin{itemize}
\item \textbf{Idiap Replay Attack} (denoted as I) \cite{chingovska2012indiap_replay}: This dataset includes 1,300 videos captured from 50 clients under two different lighting conditions. It features four types of replayed faces and one type of printed face for spoof attacks.
\item \textbf{OULU-NPU} (denoted as O) \cite{boulkenafet2017oulu_npu}: Comprising high-resolution videos, this dataset contains 3,960 spoof face videos and 990 live face videos captured from six different cameras. It includes two kinds of printed faces and two kinds of replayed faces.

\item \textbf{CASIA-MFSD} (denoted as C) \cite{zhang2012casia}: Consisting of 50 subjects, each with 12 videos, this dataset features three types of attacks: printed photo, cut photo, and video attacks.

\item \textbf{MSU-MFSD} (denoted as M) \cite{wen2015msu_mfsd}: This dataset includes 280 videos for 35 subjects recorded with different cameras. It encompasses three spoof types: one kind of printed face and two kinds of replayed faces.
\end{itemize}

 Following the pre-processing steps outlined in \cite{sun2023safas}, we utilized MTCNN \cite{zhang2017detecting} to detect faces in each frame of the videos.

\section{Proof of Theorem 1}
\label{supp:prooftheorem1}

  \textbf{Theorem 1 (\textit{Restate})}: Suppose that the loss function $\ell(\theta_{t}) = \ell(f(x;\theta_{t}), y)$ satisfies the following assumptions. (i) its gradient  $g(\theta_{t})=\nabla\ell(\theta_{t})$  is  bounded, i.e., $\lVert g(\theta_{t})\rVert \leq G \text{, } \forall t$. (ii) The stochastic gradient is L-Lipchitz, i.e., $\lVert g(\theta_{t})-g(\theta_{t}^\prime)\rVert \leq L\lVert \theta_{t}-\theta_{t}^\prime\rVert$, $\forall \theta_{t}$, $\theta_{t}^\prime$.  Let the learning rate $\eta_t$ be $ \frac{\eta_0}{\sqrt{t}}$, and and let the perturbation be proportional to the learning rate, i.e., $\rho_t=\frac{\rho}{\sqrt{t}}$, and $\gamma_t=\frac{\gamma}{\sqrt{t}}$, we have:
  \begin{align} 
      &\frac{1}{T}\Sigma^{T}_{t=1}\mathbb{E}_{\textcolor{blue}{\mathcal{S}_{i}}\sim\boldsymbol{\mathcal{S}}}\mathbb{E}_{(x,y)\sim\textcolor{blue}{\mathcal{S}_{i}}}\left[ \lVert \nabla \ell(\theta_{t})\rVert^{2}\right] \leq \mathcal{O}\left(\frac{\log T}{\sqrt{T}}\right), \text{and}  \nonumber\\
      &\frac{1}{T}\Sigma^{T}_{t=1}\mathbb{E}_{\textcolor{blue}{\mathcal{S}_{i}}\sim\boldsymbol{\mathcal{S}}}\mathbb{E}_{(x,y)\sim\textcolor{blue}{\mathcal{S}_{i}}}\left[ \lVert \nabla \ell(\theta_{t}^{\text{adv}})\rVert^{2}\right] \leq \mathcal{O}\left(\frac{\log T}{\sqrt{T}}\right), \nonumber
  \end{align}
  where  $\theta_{t}^{\text{adv}}=\theta_{t}+\hat{\epsilon_t}-\gamma_t\delta_t$, $\delta_t=\Sigma_{j=1}^{k}\nabla\ell(f(x_j^\prime;\theta_t),y_j^\prime)$ with $(x_j^\prime,y_j^\prime) \sim\mathcal{S}_{j}$. 

  For simplicity, we denote the update at step $t$ as:
  \begin{equation}
       d_t =-\eta_t g(\theta_t) -\eta_t g(\theta_{t}^{\text{adv}}).
  \end{equation}
  By L-smoothess of $\ell$ and the definition of $d_t=\theta_{t+1}-\theta_{t}$, we have: 
  \begin{align}
      &\ell(\theta_{t+1}) -\ell(\theta_{t}) \leq   
 \langle \nabla\ell(\theta_{t}), \theta_{t+1}-\theta_{t}\rangle \nonumber + \frac{L}{2}\lVert \theta_{t+1}-\theta_{t}\rVert^2 \\
 &=\langle \nabla\ell(\theta_{t}), d_t\rangle \nonumber + \frac{L}{2}\lVert d_t\rVert^2 \nonumber\\
 &=  -\eta_t\langle \nabla\ell(\theta_{t}),  g(\theta_t) + g(\theta_{t}^{\text{adv}}) \rangle \nonumber+\frac{L\eta_t^2}{2} \lVert  g(\theta_t)  + g(\theta_{t}^{\text{adv}}) \rVert^2 \nonumber \\
 &= -\eta_t\langle \nabla\ell(\theta_{t}), \nabla\ell(\theta_{t}) + g(\theta_{t}^{\text{adv}}) \rangle  +\frac{L\eta_t^2}{2} \lVert  g(\theta_t)  + g(\theta_{t}^{\text{adv}}) \rVert^2 \nonumber \\
 &= -\eta_t\langle \nabla\ell(\theta_{t}), \nabla\ell(\theta_{t}) + \nabla\ell(\theta_{t})- \nabla\ell(\theta_{t}) + g(\theta_{t}^{\text{adv}}) \rangle  \nonumber \\
 &+\frac{L\eta_t^2}{2} \lVert  g(\theta_t)  + g(\theta_{t}^{\text{adv}}) \rVert^2 \nonumber \\
  &\leq -2\eta_t\lVert  \nabla\ell(\theta_{t}) \rVert^2 - \eta_t\langle \nabla\ell(\theta_{t}),   g(\theta_{t}^{\text{adv}}) -g(t)  \rangle + L\eta_t^2 G^2 \nonumber
  \end{align}

Taking expectation on both sides, and let $\mathbb{E}_{ \substack{(x,y)\sim\textcolor{blue}{\mathcal{S}_{i}}\\ \textcolor{blue}{\mathcal{S}_{i}}\sim\boldsymbol{\mathcal{S}} }}
=\mathbb{E}_{\small \textcolor{blue}{\mathcal{S}_{i}}\sim\boldsymbol{\mathcal{S}}}\mathbb{E}_{(x,y)\sim\textcolor{blue}{\mathcal{S}_{i}}}$, we have:
  \begin{align}
      &\mathbb{E}_{ \substack{(x,y)\sim\textcolor{blue}{\mathcal{S}_{i}}\\ \textcolor{blue}{\mathcal{S}_{i}}\sim\boldsymbol{\mathcal{S}} }}\left[\ell(\theta_{t+1}) -\ell(\theta_{t})  \right ] \leq    -2\eta_t\mathbb{E}_{ \substack{(x,y)\sim\textcolor{blue}{\mathcal{S}_{i}}\\ \textcolor{blue}{\mathcal{S}_{i}}\sim\boldsymbol{\mathcal{S}} }}\left[\lVert  \nabla\ell(\theta_{t}) \rVert^2 \right ] \nonumber \\
      &+\eta_t\mathbb{E}_{ \substack{(x,y)\sim\textcolor{blue}{\mathcal{S}_{i}}\\ \textcolor{blue}{\mathcal{S}_{i}}\sim\boldsymbol{\mathcal{S}} }}\left[\langle \nabla\ell(\theta_{t}), g(t)- g(\theta_{t}^{\text{adv}})   \rangle \right ] + L\eta_t^2 G^2. \label{eqn:bound_normal}
  \end{align}

  Here we need to bound the term $\mathbb{E}_{ \substack{(x,y)\sim\textcolor{blue}{\mathcal{S}_{i}}\\ \textcolor{blue}{\mathcal{S}_{i}}\sim\boldsymbol{\mathcal{S}} }}\left[\langle \nabla\ell(\theta_{t}), g(t)- g(\theta_{t}^{\text{adv}})   \rangle \right ]$. We have:
  \begin{align}
      &\mathbb{E}_{ \substack{(x,y)\sim\textcolor{blue}{\mathcal{S}_{i}}\\ \textcolor{blue}{\mathcal{S}_{i}}\sim\boldsymbol{\mathcal{S}} }}\left[\langle \nabla\ell(\theta_{t}), g(t)- g(\theta_{t}^{\text{adv}})   \rangle \right ] \nonumber \\
      &\leq \mathbb{E}_{ \substack{(x,y)\sim\textcolor{blue}{\mathcal{S}_{i}}\\ \textcolor{blue}{\mathcal{S}_{i}}\sim\boldsymbol{\mathcal{S}} }}\left[\lVert\nabla\ell(\theta_{t})\rVert \cdot\lVert g(t)- g(\theta_{t}^{\text{adv}}) \rVert  \right ] \nonumber \\
      &\leq L\mathbb{E}_{ \substack{(x,y)\sim\textcolor{blue}{\mathcal{S}_{i}}\\ \textcolor{blue}{\mathcal{S}_{i}}\sim\boldsymbol{\mathcal{S}} }}\left[\lVert\nabla\ell(\theta_{t})\rVert \cdot\lVert  \theta_t- \theta_{t}^{\text{adv}} \rVert  \right ] \text{( assumption (ii))}\nonumber \\
      &=L\mathbb{E}_{ \substack{(x,y)\sim\textcolor{blue}{\mathcal{S}_{i}}\\ \textcolor{blue}{\mathcal{S}_{i}}\sim\boldsymbol{\mathcal{S}} }}\left[\lVert\nabla\ell(\theta_{t})\rVert \cdot\lVert  \hat{\epsilon_t}-\gamma_t\delta_t \rVert  \right ] \nonumber \\
      &\leq L\mathbb{E}_{ \substack{(x,y)\sim\textcolor{blue}{\mathcal{S}_{i}}\\ \textcolor{blue}{\mathcal{S}_{i}}\sim\boldsymbol{\mathcal{S}} }}\left[\lVert\nabla\ell(\theta_{t})\rVert \cdot\lVert  \hat{\epsilon_t}  \rVert  \right ] \nonumber \\
      &+ L\gamma_t\mathbb{E}_{ \substack{(x,y)\sim\textcolor{blue}{\mathcal{S}_{i}}\\ \textcolor{blue}{\mathcal{S}_{i}}\sim\boldsymbol{\mathcal{S}} }}\left[\lVert\nabla\ell(\theta_{t})\rVert \cdot\lVert  \delta_t \rVert  \right ]\nonumber \\
      &\leq L\rho_t\mathbb{E}_{ \substack{(x,y)\sim\textcolor{blue}{\mathcal{S}_{i}}\\ \textcolor{blue}{\mathcal{S}_{i}}\sim\boldsymbol{\mathcal{S}} }}\left[\lVert\nabla\ell(\theta_{t})\rVert \right ] \nonumber \\
      &+ L\gamma_t\mathbb{E}_{ \substack{(x,y)\sim\textcolor{blue}{\mathcal{S}_{i}}\\ \textcolor{blue}{\mathcal{S}_{i}}\sim\boldsymbol{\mathcal{S}} }}\left[\lVert\nabla\ell(\theta_{t})\rVert \cdot\lVert  \delta_t \rVert  \right ] \text{ ($\hat{\epsilon}_t\leq\rho_t$)}\nonumber \\
       &\leq L\rho_t\mathbb{E}_{ \substack{(x,y)\sim\textcolor{blue}{\mathcal{S}_{i}}\\ \textcolor{blue}{\mathcal{S}_{i}}\sim\boldsymbol{\mathcal{S}} }}\left[\lVert\nabla\ell(\theta_{t})\rVert \right ] + L\gamma_t kG\mathbb{E}_{ \substack{(x,y)\sim\textcolor{blue}{\mathcal{S}_{i}}\\ \textcolor{blue}{\mathcal{S}_{i}}\sim\boldsymbol{\mathcal{S}} }}\left[\lVert\nabla\ell(\theta_{t})\rVert    \right ] \nonumber \\
       &\leq L\rho_t G + L \gamma_t k G^2 \text{ (assumption (i))}.\label{eqn:bound_dot_grad_sub_grad}
  \end{align}

  Replace Equation \ref{eqn:bound_dot_grad_sub_grad} into Equation \ref{eqn:bound_normal} we obtain:
  \begin{align}
       &\mathbb{E}_{ \substack{(x,y)\sim\textcolor{blue}{\mathcal{S}_{i}}\\ \textcolor{blue}{\mathcal{S}_{i}}\sim\boldsymbol{\mathcal{S}} }}\left[\ell(\theta_{t+1}) -\ell(\theta_{t})  \right ] \leq    -2\eta_t\mathbb{E}_{ \substack{(x,y)\sim\textcolor{blue}{\mathcal{S}_{i}}\\ \textcolor{blue}{\mathcal{S}_{i}}\sim\boldsymbol{\mathcal{S}} }}\left[\lVert  \nabla\ell(\theta_{t}) \rVert^2 \right ] \nonumber \\
      &+L\rho_t G + L \gamma k G^2    + L\eta_t^2 G^2. 
  \end{align}
  Re-arrange the above formula, we have:
  \begin{align}
       & 2\eta_t\mathbb{E}_{ \substack{(x,y)\sim\textcolor{blue}{\mathcal{S}_{i}}\\ \textcolor{blue}{\mathcal{S}_{i}}\sim\boldsymbol{\mathcal{S}} }}\left[\lVert  \nabla\ell(\theta_{t}) \rVert^2 \right ] \leq \mathbb{E}_{ \substack{(x,y)\sim\textcolor{blue}{\mathcal{S}_{i}}\\ \textcolor{blue}{\mathcal{S}_{i}}\sim\boldsymbol{\mathcal{S}} }}\left[\ell(\theta_{t})-\ell(\theta_{t+1})  \right ]  \nonumber \\
      &+L\rho_t G + L \gamma_t k G^2    + L\eta_t^2 G^2. 
  \end{align}
  Perform telescope sum and taking expectation on each step we have:
   \begin{align}
       & 2\Sigma_{t=1}^{T}\eta_t\mathbb{E}_{ \substack{(x,y)\sim\textcolor{blue}{\mathcal{S}_{i}}\\ \textcolor{blue}{\mathcal{S}_{i}}\sim\boldsymbol{\mathcal{S}} }}\left[\lVert  \nabla\ell(\theta_{t}) \rVert^2 \right ] \leq \mathbb{E}_{ \substack{(x,y)\sim\textcolor{blue}{\mathcal{S}_{i}}\\ \textcolor{blue}{\mathcal{S}_{i}}\sim\boldsymbol{\mathcal{S}} }}\left[\ell(\theta_{0})-\ell(\theta_{T})  \right ]  \nonumber \\
      &+LG\Sigma_{t=1}^{T}\rho_t +  L  k G^2 \Sigma_{t=1}^{T}\gamma_t    + LG^2\Sigma_{t=1}^{T}\eta_t^2. \label{eqn:bound_befor_schedule} 
  \end{align}
  Note that our schedules are $\eta_t \frac{\eta_0}{\sqrt{t}}$ $\rho_t=\frac{\rho}{\sqrt{t}}$, and $\gamma_t=\frac{\gamma}{\sqrt{t}}$ then we have:
  \begin{align}
       &\frac{2\eta_0}{\sqrt{T}}\Sigma_{t=1}^{T} \mathbb{E}_{ \substack{(x,y)\sim\textcolor{blue}{\mathcal{S}_{i}}\\ \textcolor{blue}{\mathcal{S}_{i}}\sim\boldsymbol{\mathcal{S}} }}\left[\lVert  \nabla\ell(\theta_{t}) \rVert^2 \right ] \leq \text{LHS(\ref{eqn:bound_befor_schedule})} \leq \text{RHS(\ref{eqn:bound_befor_schedule})} \nonumber \\
       &\leq \ell(\theta_0) - \ell_{\text{min}} + LG\rho\Sigma_{t=1}^{T} \frac{1}{\sqrt{t}} + L  k G^2 \gamma\Sigma_{t=1}^{T} \frac{1}{\sqrt{t}}  \nonumber\\ &+ LG^2 \eta_0^2 \Sigma_{t=1}^{T} \frac{1}{t} \nonumber\\
       &\leq \ell(\theta_0) - \ell_{\text{min}} + LG\rho (2\sqrt{T}-1) + L  k G^2 \gamma  (2\sqrt{T}-1) \nonumber\\ &+ LG^2 \eta_0^2(1+\log(T)).
  \end{align}
  Hence, 
  \begin{align}
      \frac{1}{T}\Sigma_{t=1}^{T} \mathbb{E}_{ \substack{(x,y)\sim\textcolor{blue}{\mathcal{S}_{i}}\\ \textcolor{blue}{\mathcal{S}_{i}}\sim\boldsymbol{\mathcal{S}} }}\left[\lVert  \nabla\ell(\theta_{t}) \rVert^2 \right ] &\leq C_0+ \frac{C_1}{\sqrt{T}} +C_2\frac{\log T}{\sqrt{T}}\nonumber \\
      &=\mathcal{O}\left(\frac{\log T}{\sqrt{T}}\right)
  \end{align}
  where $C_0$, $C_1$, $C_2$ are some constants.

  For the second part of the Theorem, we have that :
  \begin{align}
      &\mathbb{E}_{ \substack{(x,y)\sim\textcolor{blue}{\mathcal{S}_{i}}\\ \textcolor{blue}{\mathcal{S}_{i}}\sim\boldsymbol{\mathcal{S}} }}\left[ \lVert \nabla \ell(\theta_{t}^{\text{adv}})\rVert_{2}^{2}\right] \nonumber \\
      &=\mathbb{E}_{ \substack{(x,y)\sim\textcolor{blue}{\mathcal{S}_{i}}\\ \textcolor{blue}{\mathcal{S}_{i}}\sim\boldsymbol{\mathcal{S}} }}\left[ \lVert \nabla \ell(\theta_{t})+ \nabla \ell(\theta_{t}^{\text{adv}}) - \nabla \ell(\theta_{t})\rVert^{2}\right] \nonumber \\
      &\leq 2\mathbb{E}_{ \substack{(x,y)\sim\textcolor{blue}{\mathcal{S}_{i}}\\ \textcolor{blue}{\mathcal{S}_{i}}\sim\boldsymbol{\mathcal{S}} }}\left[ \lVert \nabla \ell(\theta_{t})\rVert^{2}\right] \nonumber \\
      &+ 2\mathbb{E}_{ \substack{(x,y)\sim\textcolor{blue}{\mathcal{S}_{i}}\\ \textcolor{blue}{\mathcal{S}_{i}}\sim\boldsymbol{\mathcal{S}} }}\left[ \lVert  \nabla \ell(\theta_{t}^{\text{adv}}) - \nabla \ell(\theta_{t})\rVert^{2}\right] \nonumber \\
      &\leq 2\mathbb{E}_{ \substack{(x,y)\sim\textcolor{blue}{\mathcal{S}_{i}}\\ \textcolor{blue}{\mathcal{S}_{i}}\sim\boldsymbol{\mathcal{S}} }}\left[ \lVert \nabla \ell(\theta_{t})\rVert^{2}\right] +2\mathbb{E}_{ \substack{(x,y)\sim\textcolor{blue}{\mathcal{S}_{i}}\\ \textcolor{blue}{\mathcal{S}_{i}}\sim\boldsymbol{\mathcal{S}} }}\left[\lVert g(\theta_t^{\text{adv}})-g(\theta_t)\rVert\right]\nonumber \\
      &\leq 2\mathbb{E}_{ \substack{(x,y)\sim\textcolor{blue}{\mathcal{S}_{i}}\\ \textcolor{blue}{\mathcal{S}_{i}}\sim\boldsymbol{\mathcal{S}} }}\left[ \lVert \nabla \ell(\theta_{t})\rVert^{2}\right] \nonumber \\
      &+2L^2 \mathbb{E}_{ \substack{(x,y)\sim\textcolor{blue}{\mathcal{S}_{i}}\\ \textcolor{blue}{\mathcal{S}_{i}}\sim\boldsymbol{\mathcal{S}} }}\left[\lVert \theta_t^{\text{adv}}-\theta_t\rVert^{2}\right] \text{ (assumption (ii))}\nonumber \\
&\leq 2\mathbb{E}_{ \substack{(x,y)\sim\textcolor{blue}{\mathcal{S}_{i}}\\ \textcolor{blue}{\mathcal{S}_{i}}\sim\boldsymbol{\mathcal{S}} }}\left[ \lVert \nabla \ell(\theta_{t})\rVert^{2}\right] +2L^2 \mathbb{E}_{ \substack{(x,y)\sim\textcolor{blue}{\mathcal{S}_{i}}\\ \textcolor{blue}{\mathcal{S}_{i}}\sim\boldsymbol{\mathcal{S}} }}\left[\lVert  \hat{\epsilon_t}-\gamma_t\delta_t \rVert^{2}\right]\nonumber \\
&\leq 2\mathbb{E}_{ \substack{(x,y)\sim\textcolor{blue}{\mathcal{S}_{i}}\\ \textcolor{blue}{\mathcal{S}_{i}}\sim\boldsymbol{\mathcal{S}} }}\left[ \lVert \nabla \ell(\theta_{t})\rVert^{2}\right] +2L^2 \left( \rho_t^2 + \gamma_t^2k^2G^2\right)
  \end{align}

  Sum over $t$ and average, then we have:
  \begin{align}
      \frac{1}{T}\Sigma_{t=1}^{T}&\mathbb{E}_{ \substack{(x,y)\sim\textcolor{blue}{\mathcal{S}_{i}}\\ \textcolor{blue}{\mathcal{S}_{i}}\sim\boldsymbol{\mathcal{S}} }}\left[ \lVert \nabla \ell(\theta_{t}^{\text{adv}})\rVert_{2}^{2}\right]
      \nonumber \\
      \leq& 2\left( C_0+ \frac{C_1}{\sqrt{T}} +C_2\frac{\log T}{\sqrt{T}} \right)\nonumber \\
      +&2L^2\frac{1+\log T}{T}(\rho_0^2 + \gamma_0^2k^2G^2).
  \end{align}

  Therefor,
  \begin{align}
  \frac{1}{T}\Sigma_{t=1}^{T}\mathbb{E}_{ \substack{(x,y)\sim\textcolor{blue}{\mathcal{S}_{i}}\\ \textcolor{blue}{\mathcal{S}_{i}}\sim\boldsymbol{\mathcal{S}} }}\left[ \lVert \nabla \ell(\theta_{t}^{\text{adv}})\rVert_{2}^{2}\right] &\leq C_3+ \frac{C_4}{\sqrt{T}} +C_5\frac{\log T}{\sqrt{T}}\nonumber \\
      &=\mathcal{O}\left(\frac{\log T}{\sqrt{T}}\right)
  \end{align}
  where $C_3$, $C_4$, $C_5$ are some constants.
\section{More Empirical Experiment}
\label{sec:supp_expe}
\textbf{Details of our Implementation.}
We observe in Alg. \ref{alg:gac_fas} that random sampling in each iteration does not necessarily include images from all source domains. Specifically, the algorithm functions effectively even if, during certain iterations, the images in a minibatch belong to only 2, or even a single, domain. However, this issue can be mitigated by designing a balanced sampler. Regarding the training process, the hyperparameters are detailed precisely in Table \ref{tab:hyper_set}.

\begin{table}[!h]
    \centering
    \resizebox{0.48\textwidth}{!}{%
    \begin{tabular}{l|c|c|c|c|c}
         \toprule 
          & lr. step & FC lr. scale & Logit scale & Weight decay & Epochs\\
         \hline\hline
         {{ICM {$\rightarrow$} O}} &40   &10  &12  &$1\times 10^{-4}$ &150\\
         {{OMI {$\rightarrow$} C}} &40   &1   &16  &$5\times 10^{-4}$ &80\\
         {{OCM {$\rightarrow$} I}}  &40   &10  &32  &$6 \times 10^{-4}$ &50\\
         {{OCI {$\rightarrow$} M}}  &5   &10  &12  &$6\times 10^{-4}$ &20\\
         \bottomrule
    \end{tabular}
    }
    \caption{Hyper-parameter settings in our experiment.}
    \label{tab:hyper_set}
\end{table}

\begin{figure*}[t!]
\centering
\includegraphics[width=15.0cm]{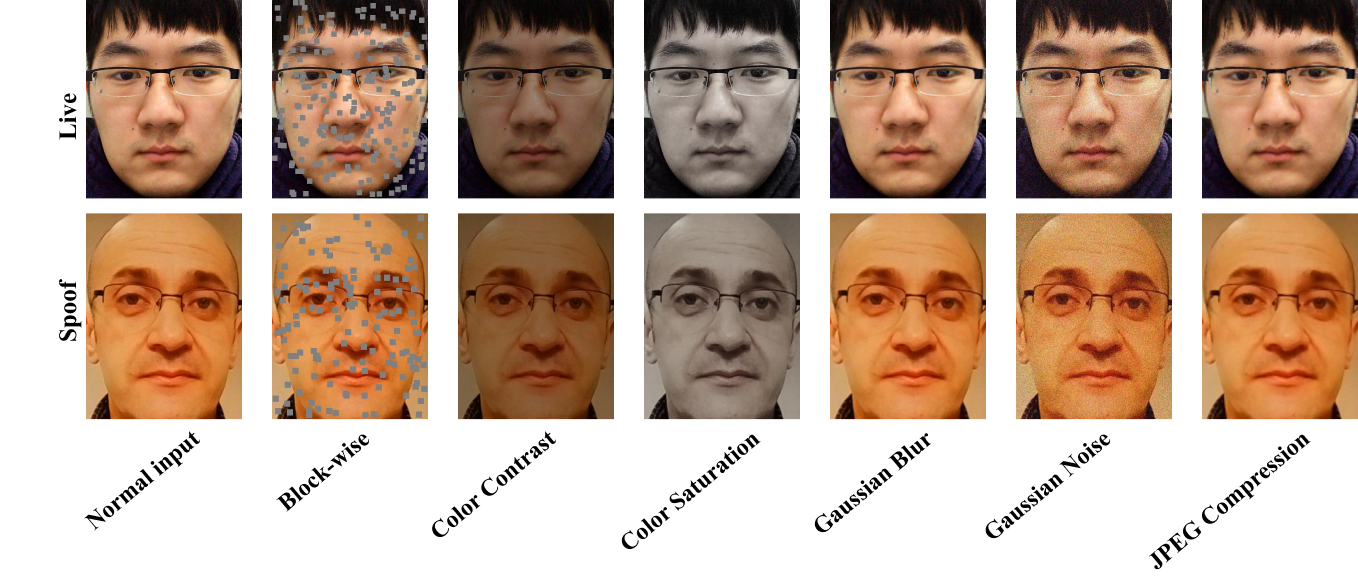}
\vspace{-8pt}
\caption{Illustration of six corruption types applied on live and spoof faces in OULU-NPU dataset.
   }
\label{fig:illus_distortion}
\end{figure*}

\textbf{Robustness to Unseen Corruptions}.
In assessing the generalization capabilities of a FAS detector, it is crucial to evaluate its robustness against various types of input corruptions, a topic extensively explored in prior works \cite{le2023quality, jiang2020deeperforensics}. Adopting the experimental settings from \cite{jiang2020deeperforensics}, we examine the detector's performance under six common image corruptions: saturation, contrast, block-wise distortion, white Gaussian noise, blurring, and JPEG compression, each with five levels of severity. In Figure \ref{fig:illus_distortion}, we showcase examples of live and spoof faces affected by \textit{six} types of image corruption techniques~\cite{jiang2020deeperforensics}, each applied with a severity level of 3. While these represent digital corruptions, they are still pertinent for assessing the resilience of spoof face detectors.

\begin{figure*}[t!]
\centering
\includegraphics[width=16.0cm]{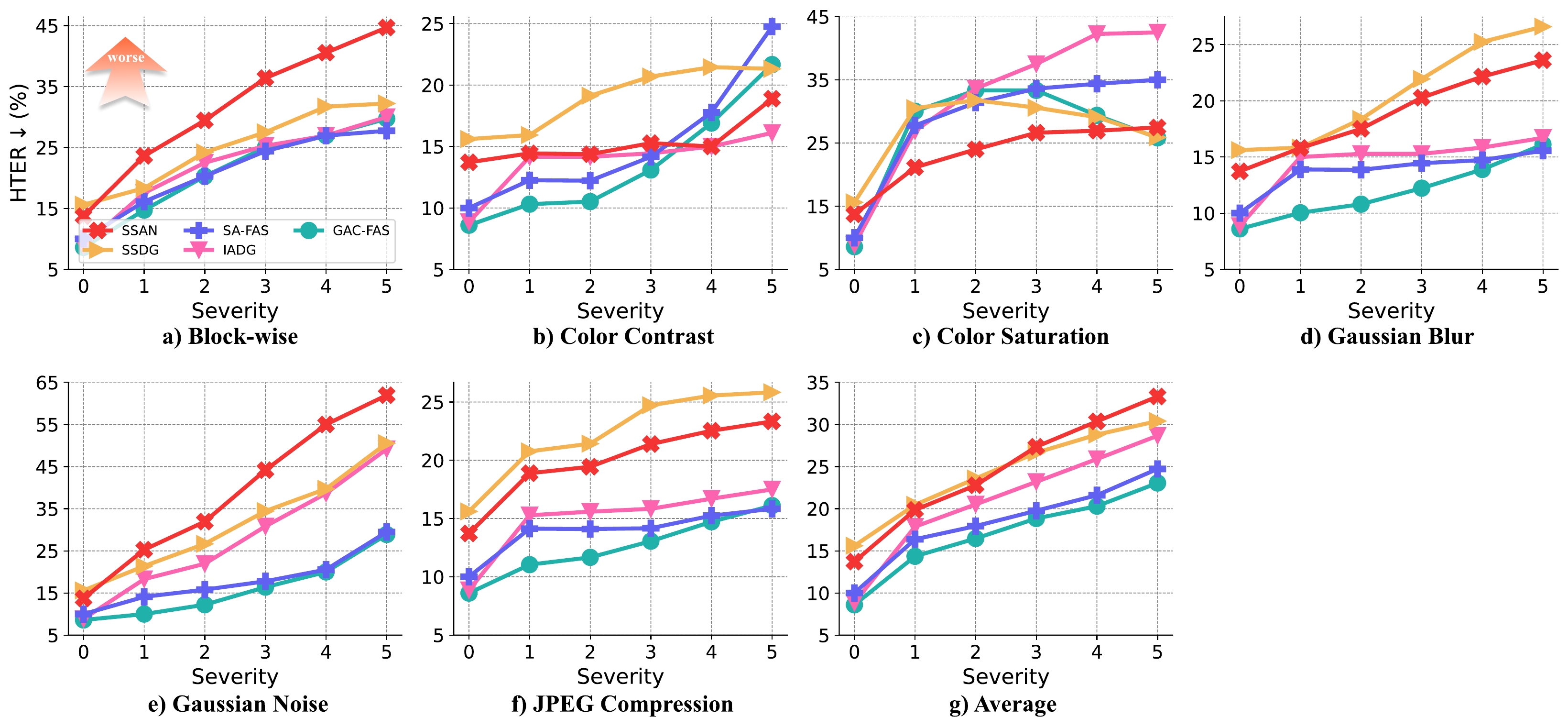}
\caption{HTER performance (\%) of DG spoof detectors under various image corruptions with different severity levels~\cite{jiang2020deeperforensics}. The experiment are conducted on ICM$\rightarrow$O with the corruptions are applied on OULU-NPU dataset.
   }
\label{fig:detail_distortion}
\vspace{-8pt}
\end{figure*}

We compare our method  with 4 baselines: SSAN~\cite{wang2022ssan}, SSDG~\cite{liu2023udgfasssdg}, SA-FAS~\cite{sun2023safas}, and IADG~\cite{zhou2023iadg}. These comparisons are based on the available official models. The results are demonstrated in Fig. \ref{fig:detail_distortion}.
Our method consistently exhibits robustness across varying severity levels, as indicated by its lower HTER performance on average.

\section{Limitations and Future Works}
\label{sec:supp_limit}
While our proposed method has achieved SoTA performance across various experiments, we acknowledge two limitations in our work. First, the training dataset requires domain labels to derive ascending points, which may limit its applicability in  the  in scenarios where training data from multiple sources are combined. Second, although our method maintains comparable computational demands to other methods during the validation phase, \SystemName\ could incur higher computational costs during training when handling a  large number of domains as  the rising number of ascending points.

In our forthcoming research, we aim to reduce the number of ascending points by exploring similarities across domains. This endeavor includes developing a more efficient regularization approach to gain deeper insights into generalization updates. Notably, our proposed method, which employs a SAM-based optimizer, demonstrates parallels in generating domain-specific gradients with meta-learning techniques \cite{chen2021d2am, shao2020rfm, wang2021sda}, albeit in a contrasting direction. While meta-learning methods require additional domain-specific gradient steps and may underperform compared to our approach, the potential synergy of combining ascending vectors from our \SystemName\ with descending vectors from meta-learning promises further enhancements in domain generalization. Our future research will concentrate on investigating these synergistic possibilities.
\end{appendices}
\end{document}